\ifdefined\pdftexversion\pdfoutput=1\fi
\documentclass[10pt,twocolumn,letterpaper]{article}

\usepackage[pagenumbers]{cvpr}
\usepackage{graphicx}

\usepackage{multirow}
\usepackage{tabularx}
\usepackage{array}
\usepackage{algorithm}
\usepackage{algpseudocode}
\usepackage{float} 
\usepackage{pifont}

\newcommand{\cmark}{\ding{51}}
\newcommand{\xmark}{\ding{55}}

\newcommand{\LSTE}{\textsc{LSTE}}

\newcommand{\STATE}{\State}
\newcommand{\REQUIRE}{\Require}
\newcommand{\ENSURE}{\Ensure}
\newcommand{\RETURN}{\Return}
\newcommand{\COMMENT}[1]{\Comment{#1}}

\newcommand{\IF}{\If}

\newcommand{\ENDIF}{\EndIf}
\newcommand{\FOR}{\For}
\newcommand{\ENDFOR}{\EndFor}

\newcommand{\PROCEDURE}{\Procedure}
\newcommand{\ENDPROCEDURE}{\EndProcedure}

\definecolor{cvprblue}{rgb}{0.21,0.49,0.74}
\usepackage[pagebackref,breaklinks,colorlinks,allcolors=cvprblue]{hyperref}

\def\paperID{xxxx}
\def\confName{CVPR}
\def\confYear{2026}

\title{Learning Whole-Body Human-Humanoid Interaction \\
 from Human-Human Demonstrations}

\author{
    Wei-Jin Huang$^{1}$ \hspace{3mm} Yue-Yi Zhang$^{1}$ \hspace{3mm}  Yi-Lin Wei$^{1}$ \hspace{3mm}  Zhi-Wei Xia$^{1}$ \hspace{3mm}  Juantao Tan$^{1}$ \\
    Yuan-Ming Li$^{1}$ \hspace{3mm} Zhilin Zhao$^{1}$ \hspace{3mm} Wei-Shi Zheng$^{1,2,3,4,\dagger}$ \\
    $^1$School of Computer Science and Engineering, Sun Yat-sen University, China \\
    $^2$Peng Cheng Laboratory, China\\
    $^3$Key Laboratory of Machine Intelligence and Advanced Computing, Ministry of Education, China; \\
    $^4$Guangdong Province Key Laboratory of Information Security Technology, China\\
    {\tt\small huangwj235@mail2.sysu.edu.cn; wszheng@ieee.org }
}

\begin{document}

\twocolumn[{
\renewcommand\twocolumn[1][]{#1}
\maketitle
\vspace{-10mm}
\begin{center}
    \centering
    \includegraphics[width=\textwidth]{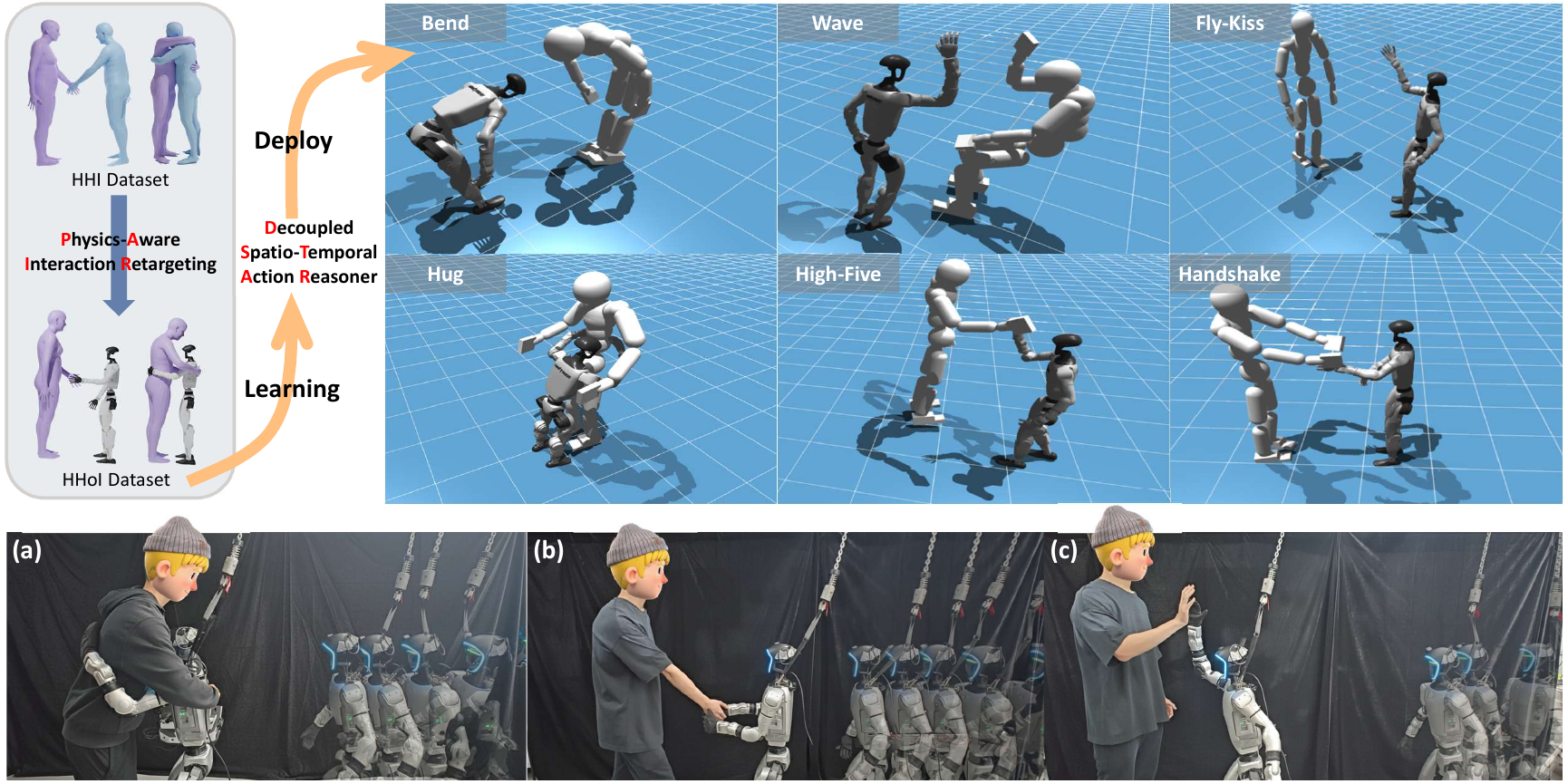}
        \vspace{-0.6cm}
        \captionof{figure}{\textbf{From HHI to HHoI with simulation and real-robot results.}
            \textbf{Left:} PAIR (Physics-Aware Interaction Retargeting) converts human--human interaction sequences into physically consistent human--humanoid (HHoI) clips by aligning morphology and explicitly preserving contact semantics via a two-stage pipeline.
            \textbf{Top (Sim):} Rollouts of the learned policy (D-STAR) in simulation, showing \emph{Bend}, \emph{Wave}, \emph{Fly-Kiss}, \emph{Hug}, \emph{High-Five}, and \emph{Handshake}, demonstrating synchronized whole-body interactions.
            \textbf{Bottom (Real, a--c):} Deployment on a Unitree G1 under a standard whole-body controller; the policy executes \emph{Hug}, \emph{Handshake}, and \emph{High-Five} selected via text commands.
        }
        \label{fig:1}
\end{center}
}]

{\let\thefootnote\relax\footnotetext{
\scriptsize
{$\dagger$}: Corresponding author.
}}

\begin{abstract}
Enabling humanoid robots to physically interact with humans is a critical frontier, but progress is hindered by the scarcity of high-quality Human-Humanoid Interaction (HHoI) data. While leveraging abundant Human-Human Interaction (HHI) data presents a scalable alternative, we first demonstrate that standard retargeting fails by breaking the essential contacts. 
We address this with \textbf{PAIR} (\emph{Physics-Aware Interaction Retargeting}), a contact-centric, two-stage pipeline that preserves contact semantics across morphology differences to generate physically consistent HHoI data.
This high-quality data, however, exposes a second failure: conventional imitation learning policies merely mimic trajectories and lack interactive understanding. We therefore introduce \textbf{D-STAR} (\emph{Decoupled Spatio-Temporal Action Reasoner}), a hierarchical policy that disentangles when to act from where to act. 
In \textbf{D-STAR}, Phase Attention (when) and a Multi-Scale Spatial module (where) are fused by the diffusion head to produce synchronized whole-body behaviors beyond mimicry.
By decoupling these reasoning streams, our model learns robust temporal phases without being distracted by spatial noise, leading to responsive, synchronized collaboration. 
We validate our framework through extensive and rigorous simulations, demonstrating significant performance gains over baseline approaches and a complete, effective pipeline for learning complex whole-body interactions from HHI data.
\end{abstract}

\section{Introduction}
\label{sec:intro}

Humanoid robots, with their human-like form, hold significant potential for seamless integration into human-centric environments, enabling intuitive collaboration and the use of human tools. While foundational capabilities such as locomotion, manipulation, and learning from human data have advanced considerably \cite{zhuang2024humanoid,tang2024humanmimic,fan2025one,he2025asap,cheng2024expressive,he2024omnih2o}, unlocking the full potential of these robots requires moving beyond single-agent proficiency. The critical next frontier lies in mastering \textbf{Human-Humanoid Interaction (HHoI)}, enabling dynamic collaboration and interaction within shared human spaces. However, achieving HHoI faces major hurdles in \textbf{Data Acquisition} and \textbf{Interaction Policy Design}. 

\begin{figure}[t]
\centering
\includegraphics[width=0.95\columnwidth]{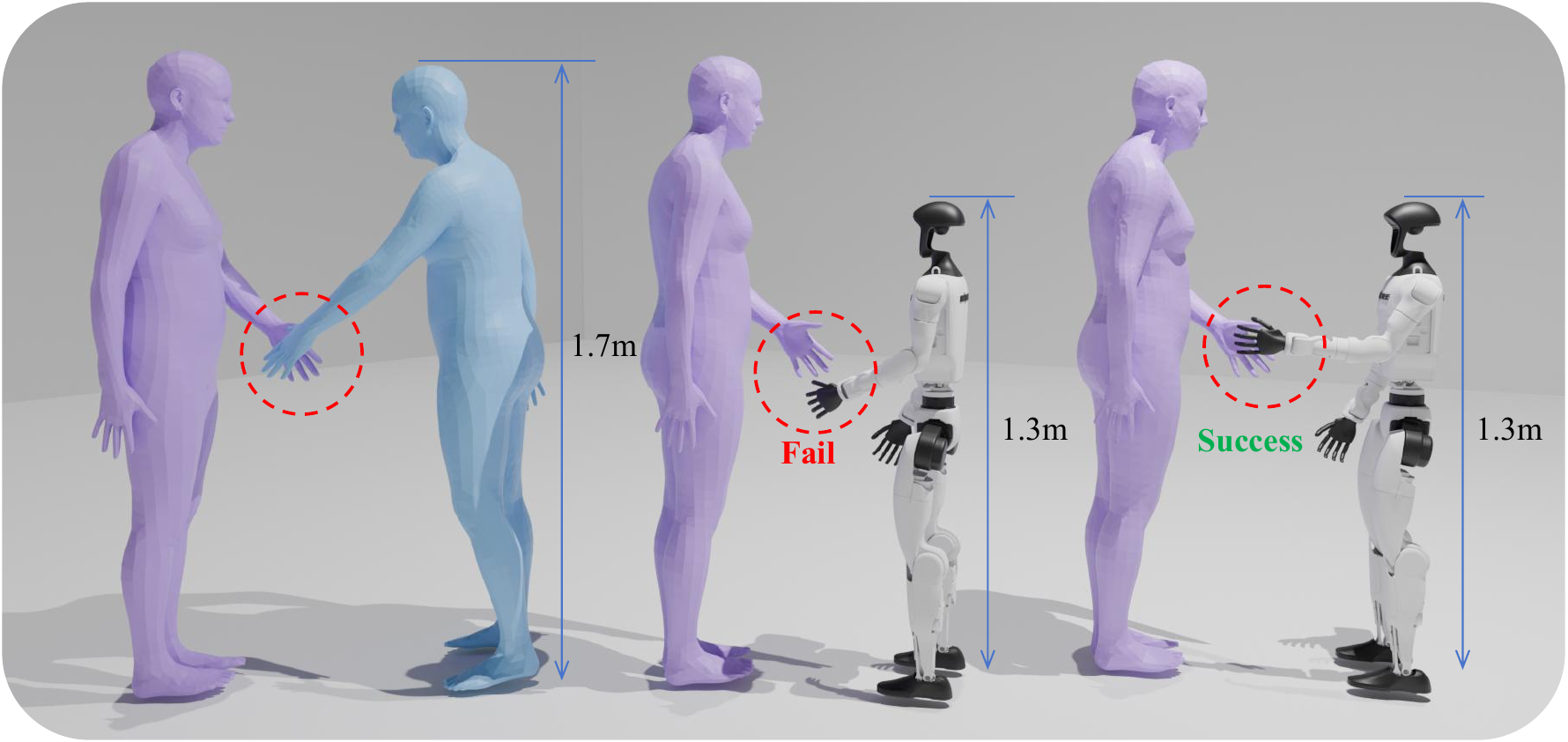} 
\vspace{-0.25cm}
\caption{\textbf{PAIR} preserves physical consistency where naive methods fail. \textbf{Left}: A source HHI handshake. \textbf{Center}: Naive retargeting breaks essential contact due to morphological disparities. \textbf{Right}: PAIR first ensures kinematic plausibility, then applies an interaction-aware objective ($\mathcal{L}_{\text{con}}$) to refine and enforce the critical physical contact.} 
\vspace{-0.5cm}
\label{fig:retarget_blender}
\end{figure}

Data scarcity is critical: while real-world data via teleoperation \cite{cardenas2024xbg} is an option, it is high-fidelity but costly, slow, potentially unsafe, and lacks diversity. A more scalable alternative is to leverage simulated data from retargeting Human-Human Interactions \cite{choi2019towards, choi2020nonparametric, yan2023imitationnet}, but current methods falter. They often prioritize semantic similarity over physical consistency, neglecting morphological differences between humans and humanoids. This can break physical contact essential for interaction (e.g., a retargeted handshake failing due to height differences, as shown in Fig.~\ref{fig:retarget_blender}), compromising realism.

Policy complexity is another major challenge. Controlling a high-DoF humanoid for safe, dynamic HHoI demands real-time management of stability, collisions, and constraints, coupled with context-aware, responsive actions. Current policies, often developed for single-agent or non-interactive settings, cannot typically interpret intent or proactively engage, which is crucial for natural interaction, rather than just mimicking motions.

To address these challenges, we propose a two-pronged solution that goes beyond simple mimicry at both the \textit{Data} and \textit{Policy} levels. 

To solve the data scarcity bottleneck, we introduce \textbf{PAIR} (\emph{Physics-Aware Interaction Retargeting}; Sec.~\ref{sec:retargeting}), a contact-centric, two-stage retargeting pipeline that preserves contact semantics and physical consistency. Instead of naively mimicking joint angles, which we show breaks critical physical contacts, PAIR preserves the interaction's physical semantics through a targeted optimization process, yielding a large-scale, physically consistent HHoI dataset.

Armed with this high-fidelity data, we find conventional imitation learning still reduces to simple trajectory mimicry, lacking true interactive understanding. It learns to replicate an ``average'' motion, lacking an understanding of relational geometry or timing for responsive interaction. To overcome this, we present \textbf{D-STAR} (\emph{Decoupled Spatio-Temporal Action Reasoner}; Sec.~\ref{sec:method}), which disentangles \textbf{when} to act (Phase Attention, \textbf{PA}) from \textbf{where} to act (Multi-Scale Spatial module, \textbf{MSS}), enabling synchronized whole-body collaboration. Fig.~\ref{fig:1} demonstrates our method's effectiveness.

In summary, our main contributions are:
\begin{itemize}
  \item \textbf{Unlocking training data from HHI to HHoI.}
  We develop \textbf{PAIR}, an interaction-aware retargeting pipeline that converts large-scale
  Human--Human Interaction (HHI) datasets into \emph{physically consistent} supervision for
  whole-body Human--Humanoid Interaction (HHoI), preserving \emph{contacts} and \emph{relative geometry}.
  \item \textbf{Decoupling when vs. where for interactive control.}
  We propose \textbf{D-STAR}, which separates \emph{temporal intent} (handled by PA)
  and \emph{spatial selection} (handled by MSS), then fuses them for synchronized whole-body actions.
  \item \textbf{Validation in simulation and real-robot.}
  We conduct extensive experiments in a high-fidelity simulator and deploy on a humanoid robot with asynchronous sensing and a standard whole-body controller.
\end{itemize}

We keep quantitative comparisons in simulation for fairness across controllers, and use real-robot trials to validate practical executability under asynchronous sensing and standard whole-body control.

\section{Related Work}
\label{sec:related_work}

\noindent\textbf{Human--Humanoid Interaction.}
Recent work advances physical interaction from several fronts: XBG~\cite{cardenas2024xbg} learns end-to-end HHoI but relies on costly teleoperation, limiting scale; transformer-based social forecasting (ECHO~\cite{mascaro2024robot}) produces open-loop trajectories without reactive control; RHINO~\cite{chen2025rhino} targets human--object interaction but is upper-body only and still requires demonstrations. \emph{Concurrently}, closed-loop whole-body directions span compliance/impedance for safer contacts~\cite{lu2025gentlehumanoid,li2025thor}, robot-side multi-task control~\cite{kalaria2025dreamcontrol}, and policies learned from human videos or unified human--robot formulations~\cite{weng2025hdmi,xu2025intermimic,pmlr-v305-qiu25a}. These lines are complementary but differ in supervision pathway and scope, typically relying on robot-side supervision or broadening the task beyond social HHI. In contrast, we pursue a scalable route to \textbf{whole-body, closed-loop} HHoI by learning from abundant HHI demonstrations and explicitly decoupling \textit{when} to act from \textit{where} to contact, enabling reactive collaboration without extensive robot-side data collection.

\noindent\textbf{Motion Retargeting for Interaction.}
Retargeting HHI to humanoids is a promising route to bridge the data gap. Classical IK/trajectory optimization~\cite{choi2019towards, gomes2019humanoid} and recent learning-based retargeting~\cite{choi2021self, yan2023imitationnet} mainly optimize kinematic similarity or style, which often fails to preserve contact semantics under morphology mismatch, yielding broken or drifting contacts in HHoI (Fig.~\ref{fig:retarget_blender}). In parallel, the graphics community has proposed geometry-/context-aware objectives for animation~\cite{jang2024geometry, cheynel2025reconform}, but these typically lack robotics-grade constraints and safety (e.g., joint limits, self/rigid-body collisions). \emph{Concurrently}, interaction-preserving retargeting with interaction meshes/Laplacian deformation~\cite{yang2025omniretarget} and evidence that retargeting quality strongly affects downstream policy performance~\cite{araujo2025gmr} have appeared. Our pipeline is tailored for HHoI and features: (i) an interaction-semantic objective via an $N{\times}N$ human--robot distance loss $L_{\text{con}}$ that prioritizes who touches what, where, and for how long over single-body style; (ii) a two-stage schedule that first secures global feasibility and then tightens contact consistency, mitigating local minima and collision artifacts; and (iii) optimization under robotics-grade constraints (limits, collisions). This closes the data--policy loop for HHoI: higher contact fidelity in retargeted data translates into stronger success of the learned interactive controller.

\section{PAIR: Physics-Aware Interaction Retargeting} 
\label{sec:retargeting}

\begin{figure}[t]
  \centering
  \includegraphics[width=0.8\linewidth]{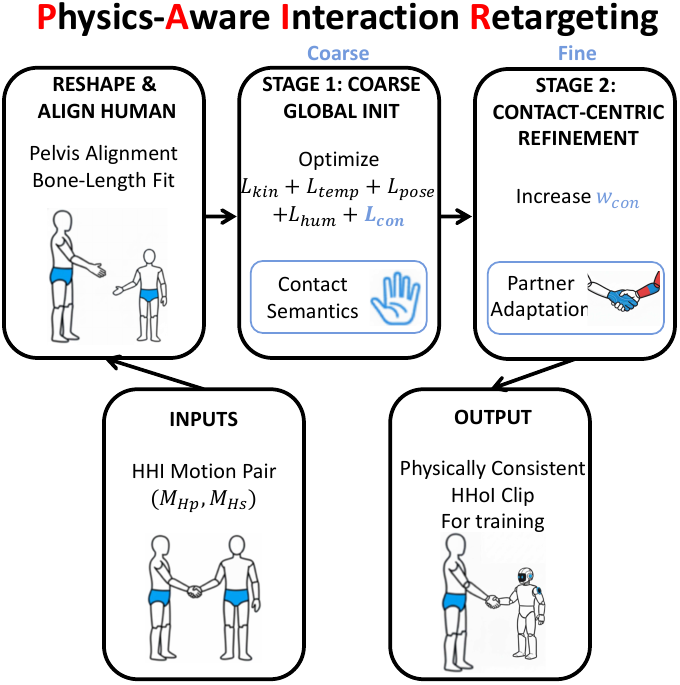}
  \vspace{-2mm}
  \caption{\textbf{PAIR} preserves contact semantics and physical consistency via a two-stage retargeting pipeline. From HHI to HHoI while retaining contact semantics across morphology differences.} 
  \label{fig:retarget-pipeline}
  \vspace{-5mm}
\end{figure}

The primary obstacle to leveraging abundant Human-Human Interaction (HHI) data is the critical failure of conventional retargeting paradigms in the context of physical interaction. The foundational principle of motion retargeting is to preserve kinematic similarity, typically by minimizing differences in joint positions or orientations. However, we contend that for interactive tasks, this very principle, when applied in isolation, leads to catastrophic failure. As visualized in Fig.~\ref{fig:retarget_blender} (Center), an approach focused solely on matching source kinematics breaks the essential physical contact that defines the interaction, a direct result of morphological disparities between the human and the humanoid. An overview is shown in Fig.~\ref{fig:retarget-pipeline}.

Sequences $\mathbf{M}_X=\{\mathbf{q}^X_t\}_{t=1}^{T}$; $\mathcal{J}_t(X,j)\in\mathbb{R}^3$ is the 3D joint position via forward kinematics from $\mathbf{q}^X_t$. $\text{Reshaped}(H_s)$ denotes the morphology-aligned human skeleton (pelvis alignment + proportional bone-length scaling + fixed joint correspondences); the full mapping appears in Suppl.~A.1, and the symbol table in Suppl.~Table~S5.

To overcome this fundamental limitation, we architect our solution as a sequence of targeted fixes. We begin by formulating the retargeting as an optimization problem over the source interaction $(M_{H_p}, M_{H_s})$ to find an optimal robot motion $M_R$ and a minimally adjusted partner motion $M'_{H_p}$. Our final objective function systematically addresses each layer of the problem:

$$
\mathcal{L}_{\text{retarget}} = w_{\text{con}}\mathcal{L}_{\text{con}} + w_{\text{kin}}\mathcal{L}_{\text{kin}} + w_{\text{hum}}\mathcal{L}_{\text{hum}} + w_{\text{reg}}\mathcal{L}_{\text{reg}}.
$$
In the following, we deconstruct how each component addresses a specific failure mode in a logical cascade.

\begin{figure*}[t]
\centering
\includegraphics[width=\textwidth]{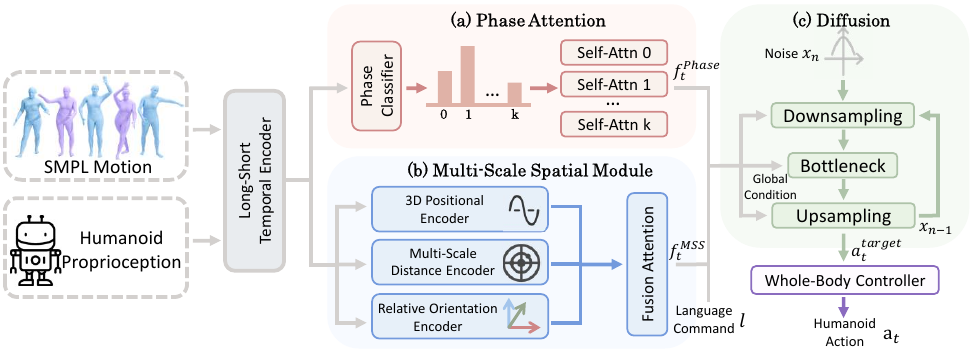} 
\vspace{-0.6cm}
\caption{Overview of \textbf{D-STAR} (Decoupled Spatio-Temporal Action Reasoner): Phase Attention (\textbf{PA}, “when to act”) and Multi-Scale Spatial module (\textbf{MSS}, “where to act”) are fused by a diffusion planning head to yield synchronized whole-body interaction beyond mimicry; a low-level Whole-Body Controller (WBC) executes the final physically plausible action.}
\label{fig:model_architecture}
\vspace{-5mm}
\end{figure*}

\subsection{From Mimicry to Interaction-Awareness}

\noindent\textbf{Kinematic Similarity Loss ($\mathcal{L}_{\text{kin}}$).} The baseline approach is to ensure that the robot's motion style resembles the original human demonstration. This is achieved by penalizing the difference between the robot's global joint positions and those of a morphologically-reshaped source human skeleton $\text{Reshaped}(H_s)$. While foundational, this is the source of the contact-breaking problem.

$$\mathcal{L}_{\text{kin}} = \frac{1}{T \cdot J_R} \sum_{t=1}^T \sum_{j=1}^{J_R} \| \mathcal{J}_t(R, j) - \mathcal{J}_t(\text{Reshaped}(H_s), j) \|_2^2.$$

\noindent\textbf{Interaction-Contact Preservation Loss ($\mathcal{L}_{\text{con}}$).}
To robustly preserve the \emph{holistic} physical semantics of interaction, we enforce consistency between the full pairwise-distance matrices computed from task keypoint sets $(K_H, K_R)$ for the original human--human interaction and the optimized human--robot interaction. Let $D^{\text{orig}}_t$ and $D^{\text{opt}}_t$ be the $N\times N$ distance matrices at time $t$; the loss is:

$$
\mathcal{L}_{\text{con}}=\frac{1}{T}\sum_{t=1}^{T}\left\| D^{\text{opt}}_{t}-D^{\text{orig}}_{t} \right\|_{F}^{2}.
$$
This matrix-level constraint maintains contact (e.g., hand-hand) and appropriate social distances beyond brittle pointwise penalties and is used consistently throughout our work.

\noindent\textbf{Human Motion Fidelity Loss ($\mathcal{L}_{\text{hum}}$).} However, strictly enforcing contact on the robot could unnaturally distort the human partner's motion, breaking the mutual plausibility of the interaction. To address this side-effect, we introduce a fidelity loss that penalizes deviations from the partner's original trajectory. This allows for small, necessary adaptations (e.g., raising an arm for a shorter robot) without creating an entirely new, unrelated action. We apply \emph{selective upper-body adaptation} (shoulders/elbows/wrists) to avoid implausible posture drift while allowing minimal, task-consistent adjustments.

$$\mathcal{L}_{\text{hum}} = \frac{1}{T} \sum_{t=1}^T \| \mathbf{p}'_{H_p, t} - \mathbf{p}_{H_p, t} \|_2^2$$ 
\noindent\textbf{Physical Plausibility Regularizers ($\mathcal{L}_{\text{reg}}$).} Finally, to ensure that the generated motions are smooth and natural, not jerky or contorted, we incorporate a standard regularization term. It consists of two components: a Temporal Coherence Loss ($\mathcal{L}_{\text{temp}}$) to penalize high-frequency jitter, and a Pose Regularization Loss ($\mathcal{L}_{\text{pose}}$) to keep joint angles within a comfortable range.

\subsection{Navigating the Optimization Landscape}
While our composite objective function now captures the full semantics of interaction, its very complexity creates a new challenge: a difficult, non-convex optimization landscape prone to poor local minima. A single-stage optimization of this function often converges to physically-implausible ``near-miss'' interactions where a handshake becomes a grasp of the air inches from the partner's hand.

To solve this, we introduce a coarse-to-fine, Two-Stage Optimization Strategy. This is not a mere heuristic, but a principled approach to guide the solver through the complex energy landscape:
\begin{itemize}
    \item \textbf{Stage 1: Global Kinematic Initialization.} We first optimize the full objective over the entire sequence with a moderate weight for the contact loss ($w_{\text{con}}$). This stage finds a globally consistent and kinematically plausible motion, effectively positioning the solver in a desirable ``basin of attraction.''
    \item \textbf{Stage 2: Contact and Stability Refinement.} Using the result from Stage 1 as a warm start, we re-run the optimization with a significantly increased contact loss weight ($w_{\text{con}}$). This aggressively corrects for subtle contact misalignments and enforces physical stability, refining the ``near-miss'' into a perfect connection.
\end{itemize}

\section{D-STAR: Decoupled Spatio-Temporal Action Reasoner}
\label{sec:method}
Conventional imitation learning fails in interactive settings because it conflates two problems: timing (\textit{when to act}) and targeting (\textit{where to act}). We address this with \textbf{D-STAR} (Decoupled Spatio-Temporal Action Reasoner), a hierarchical policy built on decoupled spatio-temporal reasoning.

\textbf{Why decoupling and Overview.}
Interactive policies must reason about \textbf{when} an event should occur and \textbf{where} contact should be established. In \textbf{D-STAR}, we decouple policy reasoning into Phase Attention (\textbf{PA}) and Multi-Scale Spatial module (\textbf{MSS}), which are later fused by the diffusion head; execution proceeds via a whole-body controller. This division isolates complementary competencies—timing vs. spatial precision—and avoids interference, while a single planning head ensures coherent actions. Fig.~\ref{fig:model_architecture} summarizes the architecture; implementation frequencies and scheduler details are deferred to the Supplement. 

Conditioned on this spatio-temporal context, a diffusion-based planning head (Sec.~\ref{sec:diff_head}) generates a high-level action target, which is then realized by a Whole-Body Controller (WBC) that ensures safe and physically plausible execution. Finally, we will outline our training procedure and losses in Sec.~\ref{sec:training}. For temporal context, the network consumes a 12-frame history (1 current + 11 past) and predicts a 5-anchor horizon covering 2\,s; the execution interface is summarized in Sec.~\ref{sec:diff_head}. The policy is conditioned by a short \emph{text command} token sequence that selects the interaction type; no task-specific weights or tuning are used at deployment.

\subsection{Observations and Temporal Encoder}
\label{sec:obs}
\textbf{D-STAR} (our policy) takes as input a history of $h$ observation frames, $\mathbf{O}_{t-h+1:t}$, each frame concatenating the humanoid's proprioceptive state $\mathbf{s}_t^R$ and the human's SMPL~\cite{loper2023smpl} joint positions $\mathbf{s}_t^H$. Human-Humanoid Interaction (HHoI) demands both long-term context to understand interaction phases and short-term cues for precise spatial targeting. To capture this duality, we employ a Long-Short Temporal Encoder (\LSTE). It uses two parallel Transformer encoders to process the observation history: a long-term encoder ($E_{long}$) over the full $h$-frame horizon for contextual understanding, and a short-term encoder ($E_{short}$) over the most recent $h'$-frame horizon ($h' < h$) for fine-grained coordination. The final temporal feature vector $\mathbf{f}_t^{temp}$ is formed by concatenating their outputs: 
\begin{equation*}
\label{eq:temporal_encoder_condensed}
\mathbf{f}_t^{temp} = \text{Concat}(E_{long}(\mathbf{O}_{t-h+1:t}), E_{short}(\mathbf{O}_{t-h'+1:t}))
\end{equation*}

\begin{table*}[t]
\centering
\small
\setlength{\tabcolsep}{4pt}
\caption{Retargeting results with ablations: physical consistency (JPE, AWD) and multi-threshold contact (Prec/Rec/F1/Acc) at $\tau\!\in\!\{0.2,0.35,0.5\}$\,m, micro-avg over frames$\times$hands. Best results in \textbf{bold}, second best \underline{underlined}; metric definitions are consolidated in Sec.~\ref{subsubsec:retarget_setup}. \textbf{PAIR} yields higher contact F1 across thresholds and the best kinematic similarity. $^\dagger$\,Re-implemented from the original paper.}
\vspace{-0.2cm}
\label{tab:retargeting_main}
\begin{tabular}{l|cc|cccc|cccc|cccc}
\toprule
& \multicolumn{2}{c|}{\textbf{Physical}} & \multicolumn{4}{c|}{\textbf{Contact@0.2m}} & \multicolumn{4}{c|}{\textbf{Contact@0.35m}} & \multicolumn{4}{c}{\textbf{Contact@0.5m}} \\
\textbf{Method} & \textbf{JPE↓} & \textbf{AWD↓} & \textbf{Prec↑} & \textbf{Rec↑} & \textbf{F1↑} & \textbf{Acc↑} & \textbf{Prec↑} & \textbf{Rec↑} & \textbf{F1↑} & \textbf{Acc↑} & \textbf{Prec↑} & \textbf{Rec↑} & \textbf{F1↑} & \textbf{Acc↑} \\
\midrule
Simple MSE & 0.188 & 0.126 & 0.860 & 0.583 & 0.695 & 0.928 & 0.841 & 0.583 & 0.688 & 0.895 & 0.870 & 0.684 & 0.765 & 0.888 \\
IK Baseline & 0.337 & 0.148 & 0.850 & 0.598 & 0.702 & 0.927 & 0.779 & 0.556 & 0.649 & 0.878 & 0.776 & 0.594 & 0.673 & 0.838 \\
Orient Base. & 0.342 & 0.204 & 0.843 & 0.384 & 0.528 & 0.802 & 0.671 & 0.306 & 0.420 & 0.731 & 0.657 & 0.252 & 0.364 & 0.642 \\
ImitationNet$^\dagger$ & \underline{0.181} & 0.195 & 0.855 & 0.421 & 0.565 & 0.831 & 0.712 & 0.388 & 0.502 & 0.794 & 0.698 & 0.313 & 0.432 & 0.715 \\
\textbf{PAIR} & \textbf{0.174} & \underline{0.095} & \textbf{0.872} & \textbf{0.679} & \textbf{0.763} & \textbf{0.941} & \textbf{0.894} & \textbf{0.794} & \textbf{0.841} & \textbf{0.954} & \textbf{0.911} & \textbf{0.812} & \textbf{0.859} & \textbf{0.932} \\
\midrule
\multicolumn{15}{c}{\textbf{Ablation Study}} \\
\midrule
w/o Human Adaptation & 0.331 & \textbf{0.091} & 0.856 & \underline{0.650} & \underline{0.738} & \underline{0.936} & \underline{0.882} & \underline{0.770} & \underline{0.823} & \underline{0.946} & \underline{0.898} & \underline{0.804} & \underline{0.848} & \underline{0.926} \\
w/o Contact Loss & 0.331 & \textbf{0.091} & 0.856 & 0.648 & \underline{0.738} & \underline{0.936} & 0.881 & 0.769 & 0.821 & \underline{0.946} & \underline{0.898} & 0.803 & \underline{0.848} & \underline{0.926} \\
w/o Two-Stage & 0.330 & 0.137 & \underline{0.864} & 0.629 & 0.728 & \underline{0.936} & 0.873 & 0.718 & 0.788 & 0.936 & 0.888 & 0.749 & 0.813 & 0.911 \\
\bottomrule
\end{tabular}
\vspace{-5mm}
\end{table*}

\begin{table}[t]
\centering
\small
\setlength{\tabcolsep}{1pt}
\caption{Retargeting results with ablations: physical plausibility (LargeAngle, AngleStd) and smoothness (Jerk mean/std). Metrics in Sec.~\ref{subsubsec:retarget_setup}; best \textbf{bold}, second \underline{underlined}.}
\label{tab:retargeting_quality}
\vspace{-0.2cm}
\newcommand{\thead}[1]{\begin{tabular}{@{}c@{}}#1\end{tabular}}
\resizebox{\columnwidth}{!}{%
\begin{tabular}{lcccc}
\toprule
& \multicolumn{2}{c}{\textbf{Physical Plausibility}} & \multicolumn{2}{c}{\textbf{Motion Smoothness}} \\
\textbf{Method} & \thead{Large Angle ↓} & \thead{Angle Std ↓} & \thead{Jerk Mean ↓} & \thead{Jerk Std ↓} \\
\midrule
Simple MSE & 0.182 & 0.405 & 0.0026 & 0.0571 \\
IK Baseline & 0.118 & \textbf{0.330} & 0.0348 & 0.0587 \\
Orient Base. & 0.298 & 0.837 & 0.0142 & 0.3025 \\
ImitationNet & 0.095 & \underline{0.341} & 0.0015 & 0.0410 \\
\textbf{PAIR} & \underline{0.093} & 0.348 & \underline{0.0008} & \underline{0.0342} \\
\midrule
\multicolumn{5}{c}{\textbf{Ablation Study}} \\
\midrule
w/o Human Adaptation (HA) & \textbf{0.092} & 0.348 & \textbf{0.0007} & \textbf{0.0340} \\
w/o Contact Loss (Lcon) & \textbf{0.092} & 0.348 & \underline{0.0008} & \textbf{0.0340} \\
w/o Two-Stage & 0.093 & 0.351 & \textbf{0.0007} & \textbf{0.0340} \\
\bottomrule
\end{tabular}
}
\vspace{-5mm}
\end{table}

\subsection{Decoupled Interaction Modules}
\label{sec:pa_mss}\label{sec:pa}\label{sec:mss}
We consolidate the two parallel interaction modules into a single section to foreground \emph{decoupled spatio-temporal reasoning}. 
\noindent \textbf{Phase Attention (PA) }focuses on \emph{when} to act by leveraging temporal phase cues and long-range context, 
while the \textbf{Multi-Scale Spatial module (MSS)} resolves \emph{where} to act via multi-scale geometric neighborhoods and relational encodings. 
Both branches produce complementary features,  $\mathbf{f}_t^{Phase}$ and $\mathbf{f}_t^{MSS}$, which jointly condition the diffusion-based planning head.

\noindent \textbf{Phase Attention (PA).}
Different phases of a Human-Humanoid Interaction (HHoI) require attending to distinct cues. For instance, during the \textit{initiation} of a handshake, the robot must focus on the human's approaching hand and torso, while in the \textit{closure} phase, precise palm alignment becomes critical. We implement a phase-aware attention mechanism that predicts the current interaction phase and uses it to weigh specialized self-attention blocks, yielding a phase-conditioned temporal feature $\mathbf{f}_t^{Phase}$. Intuitively, $\mathbf{f}_t^{Phase}$ summarizes \emph{when} an interaction event unfolds, providing a temporal gate for downstream decision-making.

\noindent\textbf{Phase taxonomy and labels.} We adopt a 3-phase scheme (Preparation, Act, Follow-up) with tri-phase segmentation per sequence, using a transition-consistency loss to mitigate boundary noise.

\noindent \textbf{Multi-Scale Spatial module (MSS).}
To determine \textit{where} to act, we encode the human-robot geometric relationship via multi-scale features: absolute positions, pairwise distances, and relative orientations of key joints. Dedicated encoders map these signals to a spatial feature $\mathbf{f}_t^{MSS}$ that captures coarse-to-fine spatial context. Intuitively, $\mathbf{f}_t^{MSS}$ encodes \emph{where} the relevant contact geometry lies, aggregating near/mid/far cues into a spatially aware representation.

Ablations (Table~\ref{tab:ablation}) show that removing either branch (\textit{w/o PA} or \textit{w/o MSS}) 
degrades performance on tasks requiring precise temporal or spatial coordination, confirming the necessity of both sides of the decoupling.

\subsection{Hierarchical Action Generation}
\label{sec:diff_head}
Conditioned on PA/MSS features and a short text command, a diffusion head predicts a high-level reference action, which a downstream WBC turns into executable joint targets. The policy operates at a lower update rate than the controller to preserve stability; intermediate waypoints and blending/scheduling are defined once in Suppl.~F. The same interface is used in simulation and on hardware; controller variants and exact I/O are listed in Suppl.~A.4.

For concision, we omit per-channel dimensions, anchor counts, and frame lengths in the main text; a complete inventory appears in Suppl. Table~S1/S3.

\subsection{Training and Objectives}
\label{sec:training}
We jointly train the temporal encoder, Phase Attention, Multi-Scale Spatial, and the diffusion planning head under a composite objective that couples diffusion action prediction with auxiliary phase-classification and geometric consistency terms. Full formulations and weights are in Suppl.~A.3, scheduling/densification details are in Suppl.~F.

\section{Experiments}
\label{sec:experiments}

Our experiments are designed to systematically validate our contributions and answer four key questions: 
\textbf{Q1:} Is our physics-aware interaction retargeting method effective? 
\textbf{Q2:} Does our hierarchical policy outperform standard baselines? 
\textbf{Q3:} Are our decoupled reasoning modules essential? 
\textbf{Q4:} Does our policy demonstrate true robustness by generalizing to unseen variations in human morphology and behavior, moving beyond simple mimicry?

\subsection{Validation of Physics-Aware Interaction Retargeting}

This section quantitatively validates that our proposed retargeting pipeline (Sec.~\ref{sec:retargeting}) generates physically consistent HHoI data while preserving interaction semantics.

\subsubsection{Setup for Retargeting Evaluation}
\label{subsubsec:retarget_setup}

We compare against four baselines: a simple MSE loss on joint positions, an IK-based tracker, an orientation-based method, and the SOTA unsupervised approach ImitationNet \cite{yan2023imitationnet}. We evaluate using 18 metrics spanning Physical Consistency, Contact Preservation, Plausibility, and Smoothness (details in the supplementary). 

\noindent \textbf{Evaluation metrics.}
We report four categories.
First, \textbf{JPE (m$\downarrow$)} is the mean per-frame L2 error between robot joints and the $\text{Reshaped}(H_s)$ joints (Sec.~\ref{sec:retargeting}).
Second, \textbf{AWD (m$\downarrow$)} is the mean absolute difference between the full N$\times$N pairwise distance matrices built from a fixed set of interaction joints (head, shoulders, elbows, wrists), see Suppl.~C.2.
Third, for \textbf{Contact (multi-$\tau$)}, we classify contact per frame for the two hand–hand pairs using optimal left/right matching at $\tau\!\in\!\{0.2,0.35,0.5\}\,\mathrm{m}$ and report micro-averaged Precision/Recall/F1/Accuracy over frames$\times$hands (protocol in Suppl.~C.2.1; no minimum-duration unless stated).
Finally, \textbf{Plausibility \& Smoothness}: \textbf{LargeAngle} is the fraction of joint-angle magnitudes $> 0.5$ rad; \textbf{AngleStd} is the standard deviation of joint-angle magnitudes over joints and frames; \textbf{Jerk Mean/Std} (m/s$^3$) are computed on 3D joint positions via third-order finite differences at the dataset frame rate (50\,Hz in our data).

\begin{table*}[t]
\centering
\small
\caption{This table compares the Success Rate (Acc \%) on six interactive tasks. Our full method significantly outperforms architectural variants and fundamental baselines, especially on complex contact-based tasks (Hug, Handshake). The failure of \textbf{Diffusion Policy}, a strong baseline using the same data and controller, highlights that simply replaying trajectories is insufficient. Ablations show that both PA and MSS modules are critical for high performance. }
\label{tab:policy_success_rate}\label{tab:ablation}
\vspace{-0.2cm}
\begin{tabular}{@{}llccccccc@{}} 
\toprule
& \textbf{Method} & \textbf{Hug} & \textbf{High-Five} & \textbf{Handshake} & \textbf{Wave} & \textbf{Bend} & \textbf{Fly-Kiss} & \textbf{Avg.} \\
\midrule

\multirow{1}{*}{Fundamental Baselines} 
& Naive Mimicry (BC only)       & 0.0          & 0.0          & 0.0          & 0.0          & 0.0          & 0.0          & 0.0 \\
\midrule

\multirow{3}{*}{Architectural Variants}
& TCN Policy                    & 53.3         & 11.1         & 12.9         & 88.9         & 86.7         & \textbf{90.9} & 49.2 \\
& Transformer Policy            & 73.3         & \textbf{44.4} & 32.3         & 92.6         & 86.7         & \textbf{90.9} & 64.3 \\
& Diffusion Policy\cite{chi2023diffusion}   & 73.3         & 3.7          & 38.7         & \textbf{96.3} & \textbf{93.3} & \textbf{90.9} & 58.7 \\
\midrule

\multirow{4}{*}{Our Method and Ablations}
& \textbf{D-STAR (Full Model)}    & \textbf{100.0} & 40.7       & \textbf{61.3} & \textbf{96.3} & \textbf{93.3} & \textbf{90.9} & \textbf{75.4} \\
& w/o PA                        & 86.7         & 22.2         & 45.2         & \textbf{96.3} & \textbf{93.3} & \textbf{90.9} & 65.9 \\
& w/o MSS                       & \textbf{100.0} & 29.6       & 32.3         & \textbf{96.3} & 86.7         & 81.8         & 64.3 \\
& w/o PA + MSS                  & 86.7         & 25.9         & 51.6         & \textbf{96.3} & \textbf{93.3} & \textbf{90.9} & 68.3 \\
\bottomrule
\end{tabular}
\vspace{-5mm}
\end{table*}

\begin{table}[t]
\centering
\small
\caption{\textbf{Robustness Matrix}: Average Success Rate (\%) under combined variations in human partner's scale and speed. The central value (bold) is our standard performance. The graceful degradation towards the edges demonstrates true robustness.}
\label{tab:robustness_matrix}
\vspace{-0.2cm}
\setlength{\tabcolsep}{5pt}
\renewcommand{\arraystretch}{1.1}
\begin{tabular}{cc|ccccc}
\toprule
& & \multicolumn{5}{c}{\textbf{Partner Motion Speed}} \\
& & \textbf{0.8x} & \textbf{0.9x} & \textbf{1.0x} & \textbf{1.1x} & \textbf{1.2x} \\
\midrule
\parbox[t]{2mm}{\multirow{5}{*}{\rotatebox[origin=c]{90}{\textbf{Partner Scale}}}}
& \textbf{0.8x} & 64.3 & 66.7 & 68.3 & 66.7 & 63.5 \\
& \textbf{0.9x} & 67.5 & 72.2 & 74.6 & 70.6 & 67.5 \\
& \textbf{1.0x} & 70.6 & \textbf{75.4} & \textbf{75.4} & 73.0 & 71.4 \\
& \textbf{1.1x} & 68.3 & 72.2 & 73.0 & 70.6 & 68.3 \\
& \textbf{1.2x} & 64.3 & 64.3 & 65.1 & 64.3 & 62.7 \\
\bottomrule
\end{tabular}
\vspace{-5mm}
\end{table}

\subsubsection{Results and Analysis}
\label{sec:retargeting_results}

As shown in Tables~\ref{tab:retargeting_main} and~\ref{tab:retargeting_quality}, existing retargeting methods break down in interactive settings: the state-of-the-art learning baseline \emph{ImitationNet} attains competitive kinematic similarity (JPE 0.181\,m) yet fails to preserve the physical essence of interaction, evidenced by a contact-preservation F1 of 0.502 at the 0.35\,m threshold. This motivates an interaction-aware retargeting approach.

Our PAIR closes this gap by maintaining physical contact, achieving a contact F1 of \textbf{0.841} (0.35\,m, micro-averaged)—a \textbf{67.5\%} relative gain over \emph{ImitationNet} and \textbf{22.2\%} over the best non-interactive baseline (\emph{Simple MSE}). These results directly address the canonical failure of naive retargeting, where morphological disparities break essential contacts (Fig.~\ref{fig:retarget_blender}).

Crucially, this does not trade off other qualities: we obtain the \textbf{best kinematic similarity} (JPE \textbf{0.174}\,m) and markedly smoother motion, with a \textbf{69\%} jerk reduction versus \emph{Simple MSE} (Table~\ref{tab:retargeting_quality}).

Ablations attribute these gains to all key components. Removing Human Adaptation (HA) or the contact loss ($L_{\mathrm{con}}$) lowers contact F1 to 0.823 and 0.821, respectively—indicating that while kinematic and regularization terms form a strong base, the final margin of physical precision requires these specialized modules. Most critically, collapsing the coarse-to-fine procedure into a single stage drops F1 to 0.788, suggesting convergence to kinematically plausible yet physically incorrect near-miss minima. Together, an explicit contact objective, flexible partner adaptation, and principled two-stage optimization act synergistically to yield the observed improvements.

Having established that \textbf{PAIR} produces high-quality, physically consistent HHoI data, we next test whether such data can train a truly interactive policy surpassing standard mimicry.

\subsection{Validation of the HHoI Policy}
Here, we evaluate the effectiveness of our decoupled spatio-temporal policy, measuring task success rates across six interaction types and comparing against architectural variants and fundamental mimicry baselines.

\subsubsection{Setup for Policy Evaluation.}
We validate our policy using the Unitree G1 humanoid in Isaac Gym~\cite{makoviychuk2021isaac} (50\,Hz control). Our full model is benchmarked against fundamental baselines (Naive Mimicry) and strong architectural variants (TCN, Transformer, and Diffusion Policies \cite{chi2023diffusion}), all trained on our retargeted data and executed in simulation with the same tracking controller (WBC-Sim) for fair comparison. 
Training details of WBC-Sim appear in Suppl.~A.3.2; the execution interface is in Suppl.~A.4.1.
For qualitative real-robot validation, we run trials using a HOMIE-based pre-trained controller \cite{ben2025homie} with the waist roll and pitch locked. Task success is computed by automated detectors. 
For Handshake, success is detected by a multi-criterion, sustained-contact rule: \(\geq 10\) consecutive frames at \(50\,\mathrm{Hz}\) with valid interpersonal distance and contact stability; exact global thresholds are consolidated in the supplementary (Suppl.~C.3; Table~S6). All HHoI policy experiments are run with a fixed random seed (42). Full implementation details, success criteria, and controller notes are in the supplementary material.

\textbf{Unified protocol.}
All methods use the same dataset splits, the same whole-body controller at 50\,Hz, and the same densification/blending and scheduling procedure. Policies produce action updates at 5\,Hz. Success criteria, thresholds, and evaluation scripts are identical across methods. Implementation details are deferred to the Supplement.

\subsubsection{Setup for Policy and Training Details}
\label{sec:setup-policy-temporal}

\textbf{Policy setup.} We use a 12-frame observation history including the current frame ($t$) at 50\,Hz, and predict 5 anchors at $t+\{0,0.5,1.0,1.5,2.0\}$ s with the policy running at 5\,Hz. Observations are represented in an ego-centric frame with root-translation offset normalization, and comprise SMPL joint positions for the human together with robot proprioception (base linear and angular velocities, gravity projection, and joint states). 
The architecture adopts a Long-Short Temporal Encoder (LSTE) coupled with two decoupled interaction modules: PA for event-phase reasoning, and MSS for proximity-aware reasoning, with near/mid/far ranges set to 0-0.3\,m, 0.3-0.8\,m, and 0.8-3.0\,m. A diffusion-based planning head outputs a 36-D reference action composed of 29 joint targets and root motion (3 translation, 4-D quaternion orientation). In simulation, only the 29 joint components are executed by the controller, while the root pose is used internally; on hardware, root motion is mapped to planar base commands. We optimize with Adam ($1\times10^{-4}$, fixed), batch size 2048, for 1000 epochs.

\subsubsection{Whole-Body Controllers (Sim and Real).}
\textbf{Simulation.} We execute predicted joint targets using an tracking controller (WBC-Sim; details in Suppl.~A.4).

\noindent \textbf{Real robot.} For safety and hardware compatibility, we use a HOMIE-based controller (waist roll/pitch locked) \cite{ben2025homie}. HOMIE uses a root-velocity interface $(v_x, v_y, \dot{\psi})$ (yaw rate); we convert our root position/rotation predictions accordingly (see Suppl.~A.4.3). Our framework is controller-agnostic and can accommodate alternatives such as GMT \cite{chen2025gmt} and Hover \cite{he2024hover} (see Suppl.~A.4.5).

\noindent \textbf{Temporal scheduler.} Policies run at \(5\,\mathrm{Hz}\) and output five anchors at \(t+\{0, 0.5, 1.0, 1.5, 2.0\}\,\mathrm{s}\). We densify to \(50\,\mathrm{Hz}\) via bilinear interpolation to form a 100-frame reference (root pose + joints) tracked by the WBC; consecutive calls are fused on overlap, with zero-order hold on delays. All rules are fixed across methods as part of the Unified protocol (details are in Suppl.~F).

\subsubsection{Experimental Results and Analysis.}

As shown in Table~\ref{tab:policy_success_rate}, \textbf{Naive Mimicry} fails (0\% mean success), verifying that a low-level controller is necessary for physical stability; merely replaying trajectories is inadequate.

Our full model, \textbf{D-STAR}, achieves a state-of-the-art 75.4\% average success, surpassing strong architectural baselines—Transformer (64.3\%) and TCN (49.2\%). The 11.1-point margin over the Transformer indicates that improvements stem from decoupled spatio-temporal reasoning rather than backbone choice. The advantage is most pronounced on contact-rich tasks such as Handshake, where success nearly doubles (61.3\% vs. 32.3\%). These gains are not incremental; they directly address the shortcomings of unified policies.

Ablations (Table~\ref{tab:policy_success_rate}, bottom) show that both PA and the MSS module are essential. Removing PA degrades temporally sensitive interactions (e.g., High-Five), while removing MSS harms spatially precise ones (e.g., Handshake). The full model—with both components—achieves the best overall performance, confirming that decoupling enables complementary temporal and spatial reasoning.

\subsubsection{Robustness to Spatio-Temporal Variations}

To rigorously test generalization rather than mimicry (Q4), we evaluate a grid of joint variations in human-partner body scale and motion speed ($0.8\times$–$1.2\times$). As shown in Table~\ref{tab:robustness_matrix}, our policy is highly robust, sustaining a performance plateau (peaking at \textbf{75.4\%}) and degrading gracefully—not catastrophically—retaining \textbf{62.7\%} success even at the most extreme corner of the grid.

The results also reveal an embodiment-aware asymmetry: the policy is more resilient with smaller partners than larger ones. We attribute this to our 1.3\,m G1 humanoid: scaling the human down reduces size mismatch, whereas scaling up places key interaction points (hands, shoulders) at unfavorable heights. Thus, the policy adapts to its own kinematic affordances rather than imitating unattainable human–human behaviors, providing quantitative evidence that our decoupled reasoning architecture learns a truly generalizable interaction model.

\begin{figure}[t]
  \centering
  \includegraphics[width=\linewidth]{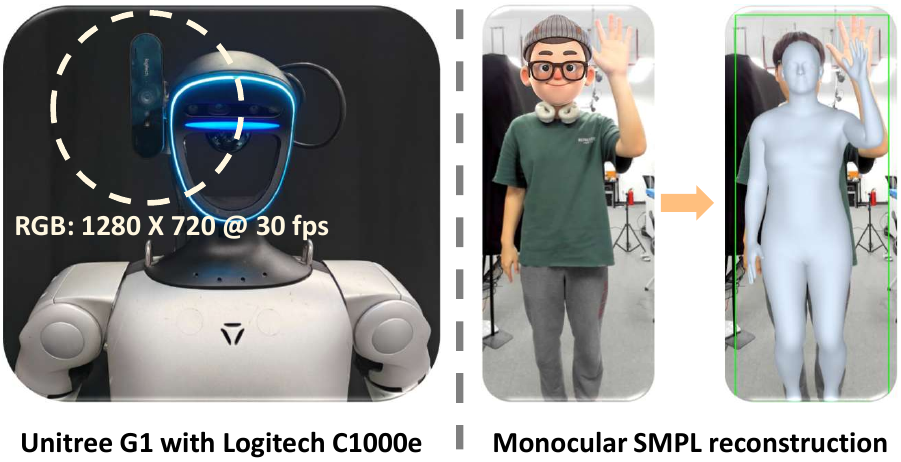}
  \vspace{-0.3cm}
  \caption{Monocular SMPL perception setup and result. (a) Unitree G1 with a Logitech C1000e RGB camera (\(1280\times720@30\) fps). (b) Input RGB frame. (c) SMPL mesh estimated from the single RGB stream (4D-Humans~\cite{goel2023humans}) and mapped to the robot base frame after extrinsic calibration.}
  \vspace{-4mm}
  \label{fig:monocular-smpl}
\end{figure}

\subsection{Sim to Real Deployment}
\label{sec:sim2real}

\textbf{Hardware \& sensing.} We deploy on a Unitree G1 (29-DoF model; practically 27-DoF with waist roll/pitch locked). A Logitech C1000e is mounted \emph{vertically} and streams 1280$\times$720@30\,fps to the control PC (Ryzen~9700X, RTX~4090). Single-RGB SMPL estimates (4D-Humans~\cite{goel2023humans}) are transformed from the camera to the robot base frame after one-time extrinsic calibration.

\noindent \textbf{Control stack and timing.} The low-level controller runs at 50\,Hz while the high-level policy runs at 5\,Hz; all remaining scheduling details mirror the simulator setup (Suppl.~F).

\noindent \textbf{Task conditioning.} The same model parameters are used; different social behaviors are selected solely via a \emph{text command} at the policy input.

\noindent \textbf{Fine-tuning for deployment.} Before deployment we perform a \emph{single, task-agnostic fine-tuning} to reduce sim-to-real gaps in pre-/post-interaction phases while keeping the core motion intact: 
(i) \textit{Act} segments are unchanged; 
(ii) \textit{Preparation} is standardized to a neutral upper-body posture---both hands naturally placed in front of the root on the left/right sides---lower body unchanged; root translation is retargeted to a milder yaw and a simpler walking pattern; the last 16 frames bilinearly transition into the first frame of the \textit{Act}; 
(iii) \textit{Follow-up} becomes stationary, and the upper body bilinearly returns over 16 frames to the same neutral hand placement. 
This retargeted dataset is used \emph{once} to fine-tune the high-level policy for real-robot execution; no per-task tuning or prompt-specific calibration is applied.

\noindent \textbf{Outcomes.} As shown in Fig.~\ref{fig:1}, the robot executes \emph{Hug, Handshake, and High-Five} (Wave and Fly-Kiss are omitted in the figure) under their respective text commands. Quantitative comparisons remain in simulation for fairness; the real-robot trials establish \emph{practical executability} under asynchronous sensing and standard whole-body control. 

\section{Conclusion}

We move beyond mimicry to learn whole-body HHoI by addressing critical failures at both the data and policy levels. First, our interaction-aware retargeting pipeline generates physically-consistent HHoI data from human demonstrations, preserving crucial contacts where naive methods fail. Building on this, our hierarchical policy uses decoupled spatio-temporal reasoning to decode interactive intent, enabling responsive collaboration rather than simple trajectory replay. Our work demonstrates a complete pipeline for learning genuine human-robot interaction, urging a shift from rote mimicry to true collaborative intelligence.

{
    \small
    \bibliographystyle{ieeenat_fullname}
    \bibliography{main}
}


\maketitlesupplementary

\renewcommand{\thetable}{S\arabic{table}}
\renewcommand{\thefigure}{S\arabic{figure}}
\renewcommand{\thesection}{\Alph{section}}
\setcounter{table}{0}
\setcounter{figure}{0}
\setcounter{section}{0}
\setcounter{algorithm}{0}

\section*{Overview}

This supplementary material is organized into six sections (A–F) for clarity and reproducibility.

\noindent\textbf{Terminology.} We refer to our retargeting pipeline as \textbf{PAIR} (\emph{Physics-Aware Interaction Retargeting}) and our policy as \textbf{D-STAR} (\emph{Decoupled Spatio-Temporal Action Reasoner}). Within D-STAR, we decouple policy reasoning into Phase Attention (\textbf{PA}) and Multi-Scale Spatial (\textbf{MSS}), which are fused by the diffusion head.

\begin{itemize}
  \item \textbf{A. Method Implementation Details} (\cref{app:method}): Retargeting pipeline (\cref{app:retargeting}), hierarchical policy (\cref{app:hsp}), training (\cref{app:train}), and low-level controllers/execution (\cref{app:llc}; root interface conversion \cref{app:rootconv}; controller modularity \cref{app:modular}).
  \item \textbf{B. Dataset Details and Analysis} (\cref{app:dataset}; built from the Inter-X dataset~\cite{xu2024inter}).
  \item \textbf{C. Experimental Protocols \& Metrics} (\cref{app:exp}): Unified notation \& symbol table (\cref{app:notation}, Table~\ref{tab:symbols}), retargeting metrics (\cref{app:retarget_eval}), policy evaluation and detection rules (\cref{app:policy_eval}).
  \item \textbf{D. Dataset Retargeting for Sim-to-Real} (\cref{app:sim2real}).
  \item \textbf{E. Observation \& Action Specification} (\cref{app:spec}; Table~\ref{tab:obs}).
  \item \textbf{F. Scheduler, Densification \& Diffusion Details} (\cref{app:schedule}; includes densification and fusion \cref{alg:densify,alg:schedule}).
\end{itemize}

\section{Method Implementation Details}
\label{sec:method_implementation}\label{app:method}

This section provides a comprehensive breakdown of our proposed framework. We first detail \textbf{PAIR} (\emph{Physics-Aware Interaction Retargeting}; Sec.~\ref{sec:sup_retargeting_details}), explaining the principled design of our objective function and optimization strategy used to generate physically consistent HHoI data.

We then dissect \textbf{D-STAR} (\emph{Decoupled Spatio-Temporal Action Reasoner}; Sec.~\ref{sec:hierarchical_policy_details}), elucidating the core principle of decoupled reasoning and its realization through specialized modules. In D-STAR, we decouple policy reasoning into \textbf{PA} and \textbf{MSS}. Finally, we specify the exact \textbf{Training Procedures and Loss Functions} (Sec.~\ref{sec:sup_training_details}) for both the high-level policy and the low-level controller.

\subsection{PAIR: Physics-Aware Interaction Retargeting} 
\label{sec:sup_retargeting_details}\label{app:retargeting}

This section provides a comprehensive breakdown of our implementation. We architect our solution as a sequence of targeted fixes, where each component of our objective function addresses a specific failure mode inherent in conventional retargeting.

\subsubsection{Objective Function Components}
We define the retargeting objective as a weighted sum of four terms that jointly encourage kinematic similarity and interaction semantics:
\begin{equation*}
\begin{split}
\mathcal{L}_{\text{retarget}} &= w_{\text{kin}}\mathcal{L}_{\text{kin}} + w_{\text{con}}\mathcal{L}_{\text{con}} + w_{\text{hum}}\mathcal{L}_{\text{hum}} + w_{\text{reg}}\mathcal{L}_{\text{reg}}, \\
\mathcal{L}_{\text{reg}} &= \alpha\,\mathcal{L}_{\text{temp}}+\beta\,\mathcal{L}_{\text{pose}}.
\end{split}
\end{equation*}
Unless otherwise stated, distances are measured in meters and joint angles in radians. The frame rate is 50\,Hz.

\paragraph{Kinematic similarity (\(\mathcal{L}_{\text{kin}}\)).}
We encourage the robot's joint-space configuration to match a morphologically reshaped source human skeleton, $\text{Reshaped}(H_s)$. The reshaping solves for SMPL~\cite{loper2023smpl} shape parameters $\boldsymbol{\beta}\in\mathbb{R}^{10}$ and a global scale $s$ that best fit the robot's bone lengths:
{\small
\begin{equation*}
\boldsymbol{\beta}^*,\, s^* \;=\; \operatorname*{\arg\min}_{\boldsymbol{\beta},\,s}
\sum_{i\in\mathcal{M}}
\left\|
\mathbf{j}_i^{R}
-
s\cdot\big(\mathbf{j}_i^{\text{SMPL}}(\boldsymbol{\beta})-\mathbf{j}_0^{\text{SMPL}}(\boldsymbol{\beta})\big)
-
\mathbf{j}_0^{R}
\right\|_2^2,
\end{equation*}}
where $\mathcal{M}$ maps robot joints to SMPL joints (correspondence listed below). We optimize this objective for 500 iterations using Adam (learning rate $0.1$).

\begin{center}
\small
\begin{tabular}{@{}lcl@{}}
\toprule
\textbf{Robot (G1)} &  & \textbf{SMPL} \\
\midrule
\texttt{pelvis}                   & $\rightarrow$ & \texttt{Pelvis} \\
\texttt{left\_hip\_pitch\_link}   & $\rightarrow$ & \texttt{L\_Hip} \\
\texttt{left\_knee\_link}         & $\rightarrow$ & \texttt{L\_Knee} \\
\texttt{left\_ankle\_roll\_link}  & $\rightarrow$ & \texttt{L\_Ankle} \\
\texttt{right\_hip\_pitch\_link}  & $\rightarrow$ & \texttt{R\_Hip} \\
\texttt{right\_knee\_link}        & $\rightarrow$ & \texttt{R\_Knee} \\
\texttt{right\_ankle\_roll\_link} & $\rightarrow$ & \texttt{R\_Ankle} \\
\texttt{left\_shoulder\_roll\_link}  & $\rightarrow$ & \texttt{L\_Shoulder} \\
\texttt{left\_elbow\_link}           & $\rightarrow$ & \texttt{L\_Elbow} \\
\texttt{left\_hand\_link}            & $\rightarrow$ & \texttt{L\_Hand} \\
\texttt{right\_shoulder\_roll\_link} & $\rightarrow$ & \texttt{R\_Shoulder} \\
\texttt{right\_elbow\_link}          & $\rightarrow$ & \texttt{R\_Elbow} \\
\texttt{right\_hand\_link}           & $\rightarrow$ & \texttt{R\_Hand} \\
\texttt{head\_link}                  & $\rightarrow$ & \texttt{Head} \\
\texttt{left\_toe\_link}             & $\rightarrow$ & \texttt{L\_Toe} \\
\texttt{right\_toe\_link}            & $\rightarrow$ & \texttt{R\_Toe} \\
\bottomrule
\end{tabular}
\end{center}

\paragraph{Interaction-contact consistency (\(\mathcal{L}_{\text{con}}\)).}
To maintain the interaction geometry, we enforce consistency between the original and optimized holistic spatial relationships. Let $\mathcal{K}_H$ and $\mathcal{K}_R$ denote selected human/robot keypoint sets (e.g., head, shoulders, elbows, wrists, hands); define the joint set $\mathcal{K}=\mathcal{K}_H\cup\mathcal{K}_R$ with $N=|\mathcal{K}|$. For time $t$, let $\mathbf{x}_t(i)\in\mathbb{R}^3$ be the position of keypoint $i\in\mathcal{K}$, and define the pairwise distance matrix
\[
[D_t]_{ij} \;=\; \|\mathbf{x}_t(i)-\mathbf{x}_t(j)\|_2,\quad D_t\in\mathbb{R}^{N\times N}.
\]
We compute $D_t^{\text{orig}}$ from the original interaction and $D_t^{\text{opt}}$ after optimization, and minimize
\begin{equation*}
\mathcal{L}_{\text{con}} \;=\; \frac{1}{T}\sum_{t=1}^{T}\left\| D_{t}^{\text{opt}}-D_{t}^{\text{orig}}\right\|_{F}^{2}.
\end{equation*}
This construction measures configuration-level geometry rather than a single contact, improving robustness to minor local deviations.

\paragraph{Human motion fidelity (\(\mathcal{L}_{\text{hum}}\)).}
To avoid unrealistic adjustments on the human partner, we penalize deviations from the original human pose while allowing selective upper-body adaptation. Let $\tilde{\mathbf{p}}_{H,t}$ be the optimized human joint positions and $\mathbf{p}_{H,t}$ the original ones. Define the upper-body index set
\(\mathcal{J}_{\text{UA}}=\{\text{shoulder},\text{elbow},\text{wrist}\}\) bilaterally. Then
\begin{equation*}
\mathcal{L}_{\text{hum}}
\;=\; \frac{1}{T}\sum_{t=1}^{T}\sum_{j\in\mathcal{J}_{\text{UA}}}\left\|\tilde{\mathbf{p}}_{H,t}(j)-\mathbf{p}_{H,t}(j)\right\|_2^2,
\end{equation*}
which enables plausible upper-limb adjustments (e.g., a slight arm lift) while preserving core posture.

\paragraph{Physical plausibility regularizers (\(\mathcal{L}_{\text{reg}}\)).}
We use standard temporal and pose regularization. Let $\mathbf{q}_t\in\mathbb{R}^{D}$ be the robot joint angles at time $t$ ($D$ DoF). We use discrete derivatives
\begin{align*}
\mathbf{v}_t &= \mathbf{q}_{t+1} - \mathbf{q}_t, \\
\mathbf{a}_t &= \mathbf{v}_{t+1} - \mathbf{v}_t = \mathbf{q}_{t+1} - 2\mathbf{q}_t + \mathbf{q}_{t-1}.
\end{align*}
The losses are
\begin{align*}
\mathcal{L}_{\text{temp}}
&= \frac{1}{T-1}\sum_{t=1}^{T-1}\|\mathbf{v}_t\|_2^2
 + \frac{w_a}{T-2}\sum_{t=1}^{T-2}\|\mathbf{a}_t\|_2^2,\\
\mathcal{L}_{\text{pose}}
&= \frac{1}{T\cdot D}\sum_{t=1}^{T}\sum_{d=1}^{D} q_{t,d}^2,
\end{align*}
where $w_a$ balances velocity and acceleration penalties (numeric values and stage-wise schedules are given in Sec.\,A.1.2).

\subsubsection{Two-Stage Optimization Strategy and Hyperparameters}
Optimizing all terms from scratch can converge to poor local minima (e.g., ``near-miss'' contacts). We therefore adopt a coarse-to-fine, two-stage schedule.

\noindent\textbf{Optimization process.}
Sequence-level retargeting is optimized for \textbf{200} iterations with Adam. The above \textbf{500} iterations refer solely to the one-time SMPL shape fitting (Sec.~A.1.1); they are not part of the per-sequence optimization.

\begin{itemize}
    \item \textbf{Stage 1 (coarse initialization)} \, (iterations 1--150, LR$=0.02$): optimize the full objective with a moderate contact weight ($w_{\text{con}}=0.25$) to obtain a kinematically feasible trajectory.
    \item \textbf{Stage 2 (contact refinement)} \, (iterations 151--200, LR$=0.005$): increase the contact weight by $10\times$ to emphasize interaction consistency ($w_{\text{con}}=2.5$), turning near-miss cases into consistent contact while keeping other weights unchanged.
\end{itemize}

\noindent\textbf{Loss weights (Stage 1).}
\begin{itemize}
    \item Kinematic similarity ($\mathcal{L}_{\text{kin}}$): $w_{\text{kin}}=1.0$
    \item Contact consistency ($\mathcal{L}_{\text{con}}$): $w_{\text{con}}=0.25$
    \item Human motion fidelity ($\mathcal{L}_{\text{hum}}$): $w_{\text{hum}}=0.25$
    \item Temporal coherence ($\mathcal{L}_{\text{temp}}$): $w_{\text{temp}}=5.0$
    \item Pose regularization ($\mathcal{L}_{\text{pose}}$): $w_{\text{pose}}=0.02$
\end{itemize}
In the notation of Sec.~A.1.1 where $\mathcal{L}_{\text{reg}}=\alpha\,\mathcal{L}_{\text{temp}}+\beta\,\mathcal{L}_{\text{pose}}$, these correspond to $w_{\text{reg}}\alpha=w_{\text{temp}}$ and $w_{\text{reg}}\beta=w_{\text{pose}}$.

\noindent\textbf{Constraints and post-processing.}
During optimization, joint angles are clamped to hardware limits $[\mathbf{q}_{\min},\,\mathbf{q}_{\max}]$. After optimization, we apply a Gaussian smoothing filter (kernel size $=5$, $\sigma=0.75$) to the joint trajectories for improved temporal smoothness.

\subsubsection{Experimental Environment}
The pipeline is implemented in PyTorch. All optimizations are performed on a server with dual AMD EPYC 7B13 CPUs, without GPU acceleration.

\subsection{D-STAR: Decoupled Spatio-Temporal Action Reasoner}
\label{sec:hierarchical_policy_details}\label{app:hsp}

\textbf{D-STAR} is architected around a core principle: \textbf{decoupling spatio-temporal reasoning}. To achieve robust and responsive interaction, an agent must solve two distinct sub-problems: inferring the temporal phase of the interaction (\textit{when} to act) and understanding the precise geometric relationship with the partner (\textit{where} to act). In D-STAR, we explicitly address this with parallel, specialized modules—Phase Attention (\textbf{PA}) and Multi-Scale Spatial (\textbf{MSS})—which are then integrated to produce a final, context-aware action. This section details the components of this architecture.

\subsubsection{Input Representation: Observation Space and Temporal Encoding}

\noindent\textbf{Observation Space Design.}
Each observation frame $\mathbf{O}_t$ concatenates human SMPL features and robot proprioception:
\begin{itemize}
    \item \textbf{Human SMPL features:} root translation $s^H_t\!\in\!\mathbb{R}^{3}$ and joint positions $j^H_t\!\in\!\mathbb{R}^{72}$ (24 joints $\times$ 3D);
    \item \textbf{Robot proprioception:} $s^R_t\!\in\!\mathbb{R}^{74}$ including base linear \& angular velocities (3+3), gravity projection (3), joint positions/velocities (29+29), root translation offset (3), and root orientation quaternion (4).
\end{itemize}
Thus $\mathbf{O}_t=\mathrm{Concat}(s^H_t, j^H_t, s^R_t)\in\mathbb{R}^{149}$.

\noindent\textbf{Long–Short Temporal Encoder (\LSTE).}
We use a dual-stream encoder to capture both long-term context and short-term reactivity:
\begin{align*}
\mathbf{f}_t^{long} &= E_{long}(\mathbf{O}_{t-h+1:t}) = \text{TranS}_{long}(\{\mathbf{O}_i\}_{i=t-h+1}^{t}),\\
\mathbf{f}_t^{short} &= E_{short}(\mathbf{O}_{t-h'+1:t}) = \text{TranS}_{short}(\{\mathbf{O}_i\}_{i=t-h'+1}^{t}).
\end{align*}
\textbf{We set $h=12$ and $h'=6$} \textit{(short stream observes only the most recent 6 frames for reactive precision, while the long stream attends to the full 12-frame buffer for phase context).} 
Each stream produces a \textbf{768-d} feature; their concatenation (\textbf{1536-d}) is linearly projected to a \textbf{256-d} temporal feature $\mathbf{f}_t^{temp}$ shared by both reasoning streams. 

\subsubsection{Temporal Reasoning: Phase Attention (PA)}

To answer \textbf{``when to act''}, PA dynamically infers the current interaction phase and emphasizes phase-relevant cues.
A phase classifier predicts a distribution $\mathbf{p}_t$ over $k=3$ phases from $\mathbf{f}_t^{temp}$:
\begin{equation*}
\mathbf{p}_t = \mathrm{Softmax}(\mathrm{MLP}_{phase}(\mathbf{f}_t^{temp})) \in \mathbb{R}^{3}.
\end{equation*}
Concurrently, $k=3$ phase-specialized self-attention experts process $\mathbf{f}_t^{temp}$. The phase-aware temporal feature is a probability-weighted mixture:
\begin{align*}
\mathbf{f}_t^{phase} &= \sum_{i=1}^{k} p_{t,i}\cdot \mathrm{SelfAttn}_i(\mathbf{Q}_t,\mathbf{K}_t,\mathbf{V}_t), \\
[\mathbf{Q}_t,\mathbf{K}_t,\mathbf{V}_t] &= \mathrm{Proj}(\mathbf{f}_t^{temp})
\end{align*}
and is projected to 256-d to match the temporal feature scale. 

\subsubsection{Spatial Reasoning: Multi-Scale Spatial (MSS)}

To answer \textbf{``where to act''}, MSS builds a scale-aware geometric understanding. We segment space into three zones (values fixed throughout):
\begin{itemize}
    \item \textbf{Near Field (0–0.3 m)}: precise contact operations,
    \item \textbf{Mid Field (0.3–0.8 m)}: coordinated gesture exchange,
    \item \textbf{Far Field (0.8–3.0 m)}: approach and planning.
\end{itemize}

\noindent\textbf{Encoders.}
A \textbf{3D positional encoder} applies sinusoidal encodings to the relative position $\mathbf{p}_{rel}$.
A \textbf{multi-scale distance encoder} processes Euclidean distance $d$ via zone-specific MLPs gated by indicator functions.
A \textbf{hierarchical orientation encoder} uses heading, projections, and rotational harmonics for approach-aware orientation.

\noindent\textbf{Fusion.}
We first concatenate encoder outputs into a spatial vector $\mathbf{f}_{spatial}$, transform it into a dynamic interaction field $\mathbf{f}_{field}$ via an MLP, and then fuse it with temporal context through attention:
\begin{equation*}
\mathbf{f}_t^{MSS} = \mathrm{FusionAttn}\!\big(\mathrm{Concat}(\mathbf{f}_{field}, \mathrm{MLP}_{proj}(\mathbf{f}_t^{temp}))\big),
\end{equation*}
followed by a projection to 128-d to obtain the final spatial feature. 

\subsubsection{Feature Fusion and Hierarchical Action Generation}

With decoupled streams providing $\mathbf{f}_t^{phase}$ and $\mathbf{f}_t^{MSS}$, we integrate them with the shared temporal context and the language command embedding to form a global conditioning vector:
\begin{equation*}
    \mathbf{c}_t = \mathrm{Concat}(\mathbf{f}_t^{temp}, \mathbf{f}_t^{phase}, \mathbf{f}_t^{MSS}, \mathbf{l}),\quad \mathbf{l}\in\mathbb{R}^{768}. 
\end{equation*}
\textbf{Dimensionalities.} $\mathbf{f}_t^{temp}\!\in\!\mathbb{R}^{256}$,\;
$\mathbf{f}_t^{phase}\!\in\!\mathbb{R}^{256}$,\;
$\mathbf{f}_t^{MSS}\!\in\!\mathbb{R}^{128}$,\;
$\mathbf{l}\!\in\!\mathbb{R}^{768}$,\;
thus $\mathbf{c}_t\!\in\!\mathbb{R}^{\mathbf{1408}}$. 

A conditional diffusion policy $\mathcal{D}$ generates a high-level target:
\begin{equation*}
\mathbf{a}_t^{target} = \mathcal{D}(\boldsymbol{\epsilon} \mid \mathbf{c}_t).
\end{equation*}
This target is executed by a pre-trained whole-body controller (WBC-Sim) to produce physically-consistent motor commands $\mathbf{a}_t$ that maintain balance and respect joint limits.
\textit{Implementation note.} In simulation we command the articulated joints (29-DoF) while using the root as an internal reference; on hardware the root component is converted to base commands via the standard base-velocity interface.

\subsection{Training Procedures and Loss Functions}
\label{sec:sup_training_details}\label{app:train}

Our framework's hierarchical nature is mirrored in our training methodology. We employ a \textbf{decoupled training strategy} that optimizes the high-level interaction policy and the low-level whole-body controller independently. This separation of concerns allows each component to be trained with the most suitable objectives and data distributions, leading to a system that is both intelligent in its interaction planning and robust in its physical execution. This section details the training procedures and objectives for each component of our hierarchy.

\subsubsection{High-Level Policy Training (Supervised Learning)}

The high-level policy, which includes the temporal encoders and our decoupled spatio-temporal reasoning modules, is trained end-to-end via supervised learning on our curated HHoI dataset.

\paragraph{Objective Function.} The policy is trained to optimize a composite loss function $\mathcal{L}_{total}$ that combines a primary action prediction objective with several auxiliary terms for regularization and phase supervision.
$$
\mathcal{L}_{total} = \lambda_{act}\mathcal{L}_{action} + \lambda_{ph}\mathcal{L}_{phase} + \lambda_{aux}\mathcal{L}_{aux}
$$
The high-level components are defined as follows:

\begin{itemize}
    \item \textbf{Action Prediction Loss ($\mathcal{L}_{action}$):} The core objective for the diffusion model is a dimension-weighted MSE loss on the predicted action $\mathbf{a}^{target}$.
    
    \item \textbf{Phase Supervision Loss ($\mathcal{L}_{phase}$):} This supervises the Phase Attention module using a cross-entropy loss for classification ($\mathcal{L}_{cls}$) and a KL divergence loss for enforcing logical phase transitions ($\mathcal{L}_{trans}$).

    \item \textbf{Auxiliary Geometric Loss ($\mathcal{L}_{aux}$):} This term encourages plausible interactive behaviors by regularizing the geometric and relational aspects of the generated motion. It includes losses for facing orientation, keypoint positioning, spatial consistency, and more.
\end{itemize}

The detailed mathematical formulations for the primary and auxiliary losses are provided below.

\noindent\textbf{Action Prediction Loss.} The primary action loss is a weighted MSE over the predicted change in root translation $\Delta \mathbf{p}$, root orientation (quaternion) $\mathbf{q}$, and joint angles $\boldsymbol{\theta}$.
\begin{align*}
    \mathcal{L}_{action} = & \quad w_{trans} \|\Delta \mathbf{p}_{pred} - \Delta \mathbf{p}_{gt}\|_2^2 \\
                     & + w_{rot} \|\mathbf{q}_{pred} - \mathbf{q}_{gt}\|_2^2 \\
                     & + w_{dof} \|\boldsymbol{\theta}_{pred} - \boldsymbol{\theta}_{gt}\|_2^2
\end{align*}
We use weights $w_{trans}=10.0$, $w_{rot}=10.0$, and $w_{dof}=1.0$.

\noindent\textbf{Auxiliary Loss Formulations.} The components of $\mathcal{L}_{phase}$ and $\mathcal{L}_{aux}$ are defined as:
\begin{itemize}
    \item \textbf{Human-Facing Orientation Loss ($\mathcal{L}_{face}$):} To ensure the robot maintains a natural orientation towards its partner, we penalize the deviation from a direct facing angle using the cosine distance between the robot's forward vector $\mathbf{v}_{fwd}^R$ and the vector to the human $\mathbf{v}_{R \to H}$.
    \begin{equation*}
        \mathcal{L}_{face} = 1 - \frac{\mathbf{v}_{fwd}^R \cdot \mathbf{v}_{R \to H}}{\|\mathbf{v}_{fwd}^R\|_2 \|\mathbf{v}_{R \to H}\|_2}
    \end{equation*}

    \item \textbf{Keypoint Position Loss ($\mathcal{L}_{pos}$):} To guide the high-level geometric structure, we apply an L2 loss on the predicted 3D positions of critical keypoints $\mathcal{K}$ (e.g., hands, head).
    \begin{equation*}
        \mathcal{L}_{pos} = \frac{1}{|\mathcal{K}|} \sum_{k \in \mathcal{K}} \| \mathbf{p}_{pred}^k - \mathbf{p}_{gt}^k \|_2^2
    \end{equation*}

    \item \textbf{Logical Phase Transition Loss ($\mathcal{L}_{trans}$):} To enforce logical temporal progression, we use a KL divergence loss between the predicted phase transition probabilities $\mathbf{P}_{t \to t+1}$ and a predefined valid transition matrix $\mathbf{T}_{valid}$.
    \begin{equation*}
        \mathcal{L}_{trans} = D_{KL}(\mathbf{T}_{valid} \parallel \mathbf{P}_{t \to t+1})
    \end{equation*}

    \item \textbf{Spatial Consistency Loss ($\mathcal{L}_{spatial}$):} To maintain the holistic geometric relationship, we penalize the difference between the predicted and ground-truth inter-keypoint distance matrices ($D_{pred}$, $D_{gt}$) using the Frobenius norm.
    \begin{equation*}
        \mathcal{L}_{spatial} = \| D_{pred} - D_{gt} \|_F^2
    \end{equation*}
\end{itemize}
The weights for these auxiliary terms are set as follows: $w_{cls}=0.0005$ (for the phase classifier), $w_{\text{phase\_KL}}=0.005$, $w_{face}=0.2$, $w_{pos}=5.0$, and $w_{spatial}=0.02$.

Table~\ref{tab:sx_obs} enumerates the oracle/student observation channels and how they are consumed during training and evaluation.

\begin{table*}[t]
\small
\centering
\caption{\textbf{WBC-Sim observation inventory and usage.} At train time, the oracle uses \emph{clean} observations without history; the student consumes \emph{noisy} observations with 10-frame history and the policy’s 36-D reference $(u_t)$. At evaluation, the fixed student is used; we retain the 10-frame history but do not inject synthetic noise. Only the WBC output (\textbf{29-D} joint-angle targets at \textbf{50\,Hz}) is actuated in simulation.}
\vspace{-2mm}
\label{tab:sx_obs}
\begin{tabular}{lcccccc}
\toprule
\multirow{2}{*}{Channel (actor)} & \multirow{2}{*}{Symbol} & \multirow{2}{*}{Dim} & Oracle & Student & Eval & History \\
 &  &  & train & train & (policy exec.) & (frames) \\
\midrule
Joint positions (target space) & $q$ & 29 & \cmark & \cmark & \cmark & 10 \\
Joint velocities               & $\dot q$ & 29 & \cmark & \cmark & \cmark & 10 \\
Base angular velocity (body)   & $\omega_b$ & 3 & \cmark & \cmark & \cmark & 10 \\
\textit{Base linear velocity (body)} & $v_b$ & 3 & \cmark & --- & --- & 0 \\ 
Gravity vector (body frame)    & $g_b$ & 3 & \cmark & \cmark & \cmark & 10 \\
Previous action (joint target) & $a_{t-1}$ & 29 & \cmark & \cmark & \cmark & 10 \\
Policy reference (from high-level) & $(u_t)$ & \textbf{36} & \xmark & \cmark & \cmark & 10 \\
\midrule
\textit{Actor total (clean vs noisy)} & & \textit{163 / 1090} & \cmark & \cmark & \cmark & \\
\textit{Critic total (clean vs noisy)} & & \textit{361 / 1291} & \cmark & \cmark & \cmark & \\
\addlinespace[0.25em]
\multicolumn{7}{l}{\footnotesize Evaluation uses the fixed \textbf{student} controller with \textbf{10-frame history} and \textbf{no synthetic observation noise}.} \\
\multicolumn{7}{l}{\footnotesize \textit{Note.} The oracle has access to privileged base linear velocity; the student excludes it.} \\ 
\bottomrule
\end{tabular}
\end{table*}

\begin{table}[t]
\ContinuedFloat
\centering
\caption{(continued): Observation and action channels. ``Used'' indicates channels that are fed to the policy; others are logging-only when present in raw data.}
\vspace{-2mm}
\label{tab:obs}
\begin{tabular}{lcc}
\toprule
Channel & Dim. & Used \\
\midrule
Human root translation & 3 & \cmark \\
SMPL joints (human) & 72 & \cmark \\
Base linear velocity & 3 & \cmark \\ 
Base angular velocity & 3 & \cmark \\
Gravity projection    & 3 & \cmark \\
Robot joint positions & 29 & \cmark \\
Robot joint velocities& 29 & \cmark \\
Root translation (offset) & 3 & \cmark \\
Root orientation (quaternion) & 4 & \cmark \\
\midrule
\textbf{Reference action (policy head)} & \textbf{36} & \textbf{—} \\
\quad Joint targets (executed) & 29 & — \\
\quad Root translation (reference) & 3 & — \\
\quad Root orientation (reference) & 4 & — \\
\bottomrule
\end{tabular}
\vspace{-5mm}
\end{table}

\paragraph{Implementation Parameters.}
The high-level policy is trained with the following configuration:
\begin{itemize}
    \item \textbf{Temporal Window:} 12-frame observation history including the current frame ($t$) at 50\,Hz; predict \textbf{5 anchors} at $t+\{0,0.5,1.0,1.5,2.0\}$\,s; anchors are densified to 50\,Hz and fused across consecutive 5\,Hz calls via bilinear interpolation.
    \item \textbf{Network Dimensions:} \LSTE long/short streams each 768-d; concatenation 1536-d with linear projection to 256-d. Phase Attention (PA) uses 256-d features; Multi-Scale Spatial (MSS) uses 128-d. Global conditioning for the diffusion head is 640-d from concatenation (256, 256, 128), with 768-d language features.
\item \textbf{Training Scale:} 1000 epochs, Adam ($1\times 10^{-4}$), batch size 2048 (stride=25), seed 42.
\item \textbf{Advanced Features:} Ego-centric processing for consistent spatial reasoning; auxiliary observations include robot proprioception and 768-d language features.
\end{itemize}

\subsubsection{Low-Level Whole-Body Controller (WBC-Sim): Oracle$\rightarrow$Student Distillation}\label{sec:wbc_distill}

\paragraph{Overview.} We train the simulation controller in two steps: an oracle teacher on clean observations and a student distilled on noisy observations with 10-frame history. Both stages share the same robot and control interface (29-D joint-angle targets at 50\,Hz), the same environment and motion-resampling interval (500 steps (50 Hz)), and the same base domain randomization; noise and DR curricula are disabled. The distilled student is the only controller used for all policy evaluations. Table~\ref{tab:wbc_distill} summarizes the teacher vs student configuration.

\paragraph{Privileged$\rightarrow$Noisy Distillation.}
We frame WBC-Sim training as privileged$\rightarrow$noisy distillation: the \emph{oracle} is optimized on clean, \emph{privileged} observations (including base linear and angular velocities), while the \emph{student} operates on noisy, realistic observations \emph{without} base linear velocity and with a 10-frame history. All evaluations use the distilled student. 

\noindent\textbf{Noise protocol.} All \emph{current-frame sensing} channels are perturbed with fixed per-channel uniform noise during student training; \emph{history stacks inherit} the perturbation from their constituent frames and receive no extra injection; the \emph{previous action} is left clean to prevent compounding errors. Table~\ref{tab:sx_noise} details the per-channel perturbations.

\paragraph{Oracle (Teacher) — Clean Observations.} Actor/critic receive noise-free raw channels (163-D / 361-D). We optimize an Upper-Body-Emphasized Tracking Reward with PPO using an MLP(512,256,128), clip 0.2, $\gamma$ 0.99, $\lambda$ 0.95, entropy 0.01, lr $1\mathrm{e}{-3}$. Base domain randomization (mass/friction/PD/push up to 1.0 m/s) is enabled; curricula are off. A frozen checkpoint is used for distillation.

\paragraph{Student — Pure Action-Matching Distillation.} Inputs are Noisy Observations with 10-Frame History (1090-D / 1291-D). We train with a pure distillation loss
\begin{equation*}
\mathcal{L}_{\text{distill}} = \left\|\mu_s\big(o_{\text{noisy}}\big) - \mu_t\big(o_{\text{clean}}\big)\right\|_2^2,
\end{equation*}
freezing the teacher and the policy std (init = 0, min = 0). Network/optimizer mirror the teacher (lr $1\mathrm{e}{-3}$). \textbf{Actions and history channels are noise-free}, while other keys use fixed per-key uniform noise (no curriculum). DR settings match the teacher.

\paragraph{Reward (key terms).} The tracking suite emphasizes upper-body targets (e.g., root\_position\_xy 3.0, body-position (upper-body) 4.0, SMPL-hands-ori 5.0) with standard stability/effort penalties and a termination cost of $-100$.

\begin{table*}[h]
\centering
\small
\caption{Teacher vs Student configuration for WBC-Sim distillation.}
\vspace{-2mm}
\label{tab:wbc_distill}
\resizebox{\linewidth}{!}{%
\begin{tabular}{@{}lllllll@{}}
\toprule
\textbf{Component} & \textbf{Obs. Set} & \textbf{Dim (act/crit)} & \textbf{Noise} & \textbf{History} & \textbf{Curric.} & \textbf{Objective} \\
\midrule
Oracle (Teacher) & Clean observations (+ base linear velocity) & 163 / 361 & Off & 0 & Off & PPO w/ tracking reward \\ 
Student (WBC-Sim) & Noisy observations with 10-frame history (\textminus{} base linear velocity) & 1090 / 1291 & On (per-key; actions \& history off) & 10 & Off & Action-mean MSE to teacher \\ 
\bottomrule
\end{tabular}
}
\end{table*}

\begin{table*}[t]
\small
\centering
\caption{\textbf{Noise model for student distillation.} We apply per-channel, zero-mean uniform noise during \emph{student training} only; the oracle uses clean observations and evaluation does not inject synthetic noise. History and action channels are not perturbed.}
\vspace{-2mm}
\label{tab:sx_noise}\label{tab:noise}
\begin{tabular}{lccc}
\toprule
Channel group & Distribution & Magnitude (symbolic) & Applied to \\
\midrule
Joint positions $q$           & $U(-\delta_q,\delta_q)$       & $\delta_q$ (const.)            & Student train \\
Joint velocities $\dot q$     & $U(-\delta_{\dot q},\delta_{\dot q})$ & $\delta_{\dot q}$ (const.)    & Student train \\
Base ang. vel. $\omega_b$     & $U(-\delta_{\omega},\delta_{\omega})$ & $\delta_{\omega}$ (const.)    & Student train \\
Gravity vector $g_b$          & $U(-\delta_{g},\delta_{g})$   & $\delta_{g}$ (const.)          & Student train \\
Prev. action $a_{t-1}$, History stacks & \multicolumn{3}{c}{\textit{no perturbation}} \\
Policy reference $u_t$ (36-D) & $U(-\delta_{u},\delta_{u})$   & $\delta_{u}$ (const.)          & Student train \\
\midrule
Curriculum & \multicolumn{3}{c}{\textit{disabled (fixed magnitudes throughout training)}} \\
Evaluation & \multicolumn{3}{c}{\textit{no synthetic noise injected}} \\
\bottomrule
\end{tabular}
\end{table*}

\subsection{Low-Level Controllers \& Execution}
\label{sec:sup_lowlevel}\label{app:llc}

This section details execution controllers in simulation and on the real robot, and outlines the interface conversion and robustness adaptations.

\subsubsection{Simulation Controller (WBC-Sim).}
We execute 29-D joint-angle targets (rad) at 50\,Hz, produced by WBC-Sim. Here, $x_t$ denotes robot proprioception (joint states, base signals, gravity, previous action, etc.). At each step, WBC-Sim takes as input (i) robot proprioception and (ii) the policy’s 36-D reference $u_t$ (29 joint targets + root translation and unit quaternion), and tracks these references under joint-limit and rate constraints. Only the 29-D joint targets are actuated in simulation; the root translation/rotation are used internally by WBC-Sim as reference signals (not directly actuated). All quantitative policy results in the main paper use the same WBC-Sim for fairness across methods. This controller corresponds to the distilled student described in Sec.~\ref{sec:wbc_distill}. For clarity, the policy observation used by the learning agent differs from the controller observation summarized here; this distinction is intentional.

\begin{equation*}
    a_t = \mathrm{WBC\text{-}Sim}(x_t, u_t),
\end{equation*}
where $u_t\in\mathbb{R}^{36}$ is the policy reference and $a_t\in\mathbb{R}^{29}$ are joint-angle targets executed at 50\,Hz.

\subsubsection{Real-World Controller (HOMIE-based, 27 DoF).}
For real-robot demos we employ a HOMIE-based pre-trained controller \cite{ben2025homie}, with the waist roll/pitch locked. HOMIE exposes a root-velocity interface $(v_x, v_y, \dot{\psi})$ (yaw rate); our planning model outputs root position/rotation, so we apply a deterministic conversion prior to execution (see Sec.~\ref{app:rootconv}). Real-robot demonstrations are qualitative; quantitative comparisons are reported in simulation using WBC-Sim.

\subsubsection{Root Interface Conversion (Position-to-Velocity Control)}
\label{app:rootconv}

Since the HOMIE\cite{ben2025homie} controller expects velocity commands $(v_{x}, v_{y}, \dot{\psi})$ while our policy predicts root position targets, we implement a closed-loop controller with active braking to ensure precise stopping. Instead of simple finite-difference conversion, we employ a distance-to-go logic based on onboard odometry. 

We issue a constant forward reference velocity $v_{ref} = 0.2$ m/s to drive the robot. During execution, we continuously monitor the residual distance, defined as the difference between the policy's cumulative predicted displacement and the actual distance traversed via odometry. To mitigate inertial drift, we implement an \textbf{active braking mechanism}: when the residual distance drops below a proximity threshold ($\delta_{dist} \le 0.15$ m) while the robot maintains significant momentum (current speed $v_{curr} \ge 0.1$ m/s), we issue a reverse velocity command ($v_{brake} = -0.2$ m/s) to rapidly decelerate and stabilize the robot at the target location.

\subsubsection{Phase-Specific Trajectory Adaptation for Stability}

To enhance execution robustness in simulation, we perform a deterministic adaptation of the reference trajectory based on the interaction phase. The \textbf{Act} phase is strictly preserved to maintain the fidelity of the learned interaction intent. For the \textbf{Preparation} and \textbf{Follow-Up} phases, we employ smoothed primitives to ensure stable transitions and prevent transient oscillations:

\begin{itemize}
    \item \textbf{Neutral Upper-Body Pose ($q_{neutral}$):} We define a stable, task-agnostic upper-body configuration used as the anchor for non-interaction phases. The joint angles (in radians) are set as follows:
    \begin{itemize}
        \item \textit{Left Arm:} Shoulder (pitch/roll/yaw) = $[0.0, 0.3, 0.0]$, Elbow = $1.0$, Wrist = $[0.0, 0.0, 0.0]$.
        \item \textit{Right Arm:} Shoulder (pitch/roll/yaw) = $[0.0, -0.3, 0.0]$, Elbow = $1.0$, Wrist = $[0.0, 0.0, 0.0]$.
    \end{itemize}

    \item \textbf{Preparation Phase:} To simulate diverse approach scenarios, we initialize the robot's root position randomly within a $1.0\,\text{m}$ radius of the original sequence's starting point, while maintaining the original orientation. The robot's root pose (position and rotation) and upper-body joints are then generated via \textit{bilinear interpolation} from this randomized state to the first frame of the Act phase, ensuring a smooth approach trajectory.

    \item \textbf{Follow-Up Phase:} Upon completion of the interaction (end of Act phase), the robot's root is kept stationary. The upper-body joints are bilinearly interpolated from the last frame of the Act phase back to $q_{neutral}$, ensuring a stable disengagement.
\end{itemize}

\subsubsection{Controller Modularity.}
\label{sec:sup_lowlevel_mod}\label{app:modular}
Our framework is controller-agnostic: WBC-Sim and the HOMIE-based controller can be replaced by alternative whole-body controllers. In particular, GMT \cite{chen2025gmt} is compatible with our action interface; we leave a systematic comparison for future work.

\begin{figure*}[t]
\centering
\includegraphics[width=\textwidth]{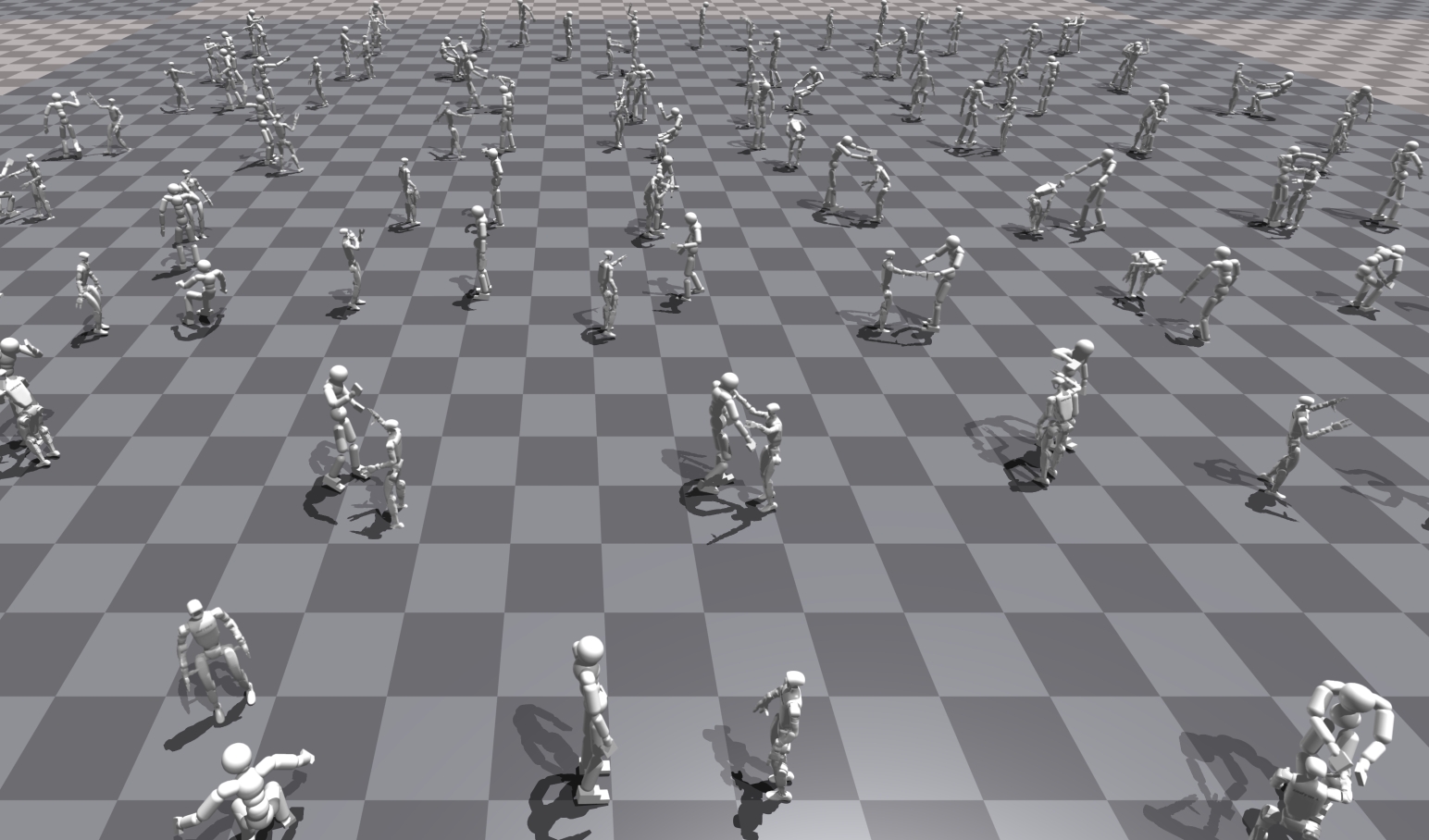} 
\caption{The foundation of our entire framework: a large-scale, high-fidelity Human-Humanoid Interaction (HHoI) dataset that we generated to solve the critical data scarcity problem. Visualized here are thousands of physically-consistent interaction pairs, the output of \textbf{PAIR}. The dataset's strategic diversity—spanning contact-rich scenarios (e.g., hugs, handshakes), dynamic gestures (e.g., high-fives), and social cues (e.g., bows)—provides the rich and complex spatio-temporal variations required to train a policy that learns to interact.} 
\label{fig:dataset}
\end{figure*}

\section{Dataset Details and Analysis}
\label{sec:dataset_details}\label{app:dataset}

In this section, we detail the construction of our Human-Humanoid Interaction (HHoI) dataset. We emphasize that this dataset is not merely a collection of motions, but a \textbf{strategically engineered resource}. The principled task selection, rigorous quality control, and, most critically, the detailed phase annotations were all designed with the explicit goal of creating a benchmark that could validate our core hypothesis: that genuine interaction requires decoupling spatial reasoning. The structure of this dataset is therefore intrinsically linked to the architecture of our policy.

\subsection{Data Source and Task Selection Rationale}

Our data foundation is the Inter-X dataset \cite{xu2024inter}, a large-scale repository of human-human interactions. To construct a comprehensive benchmark for HHoI, our task selection was not arbitrary but driven by a principled strategy to cover a taxonomy of interaction types crucial for humanoid robotics:
\begin{itemize}
    \item \textbf{Contact-Rich Interactions} (Hug, Handshake): Tasks demanding precise physical contact, stability, and management of close-proximity geometry.
  \item \textbf{Dynamic Gestural Interactions} (High-Five, Wave): Tasks requiring temporal synchronization and spatial accuracy without sustained contact.
  \item \textbf{Socially-Encoded Interactions} (Bend, Fly-Kiss): Tasks that rely on conveying social intent through whole-body gestures, often at a distance.
\end{itemize}
This curated selection of six tasks ensures our policy is evaluated across a diverse and representative range of physical and social challenges inherent to HHoI.

\subsection{Curation Pipeline and Quality Control}

Following task selection, each candidate sequence from Inter-X passed through a rigorous, multi-stage curation pipeline to ensure its suitability for training a robust HHoI policy. We enforced strict quality gates to filter out unsuitable data:
\begin{itemize}
    \item \textbf{Motion Fidelity:} We excluded samples with significant tracking errors, occlusions, or unnatural motion artifacts that would introduce noise into the training process.
    \item \textbf{Interaction Clarity:} We selected samples exhibiting clear initiation, execution, and completion phases, ensuring the full arc of an interaction was captured.
    \item \textbf{Kinematic Feasibility:} We ensured that the captured interactions occur within a spatial volume and pose range that is reasonably achievable by our target humanoid robot.
    \item \textbf{Temporal Completeness:} We retained only samples containing the complete interaction sequence, from the initial approach to the final departure.
\end{itemize}
Only sequences that passed all four gates were admitted into our final dataset, guaranteeing a high-quality foundation for our experiments.

\subsection{Phase Annotation for Interaction Understanding}

A cornerstone of our work, and a key enabler for our policy's temporal reasoning, is the detailed manual annotation of interaction phases. Recognizing that interactions are not monolithic but possess a distinct temporal structure, we manually segmented each curated sequence into three semantically meaningful phases:
\begin{itemize}
    \item \textbf{Preparation Phase:} The initial approach and positioning, where agents navigate and adjust their orientation towards each other.
  \item \textbf{Act Phase:} The core execution of the interaction, including contact establishment (e.g., Hug, Handshake), gesture performance (e.g., Wave, High-Five), or expression delivery (e.g., Bend, Fly-Kiss).
    \item \textbf{Follow-up Phase:} The completion and disengagement, involving contact release, returning to a neutral posture, and spatial separation.
\end{itemize}
This annotation process represents a significant human effort and provides the critical supervision signal for our Phase-Aware Attention module (Main.~Sec.~4.2). It directly enables our policy to learn \textit{when} to act—a capability that conventional imitation learning approaches inherently lack.

\subsection{Final Dataset Statistics and Properties}

The output of this comprehensive pipeline—from principled selection to manual annotation—is the HHoI dataset used throughout our experiments. The key statistics, which reflect the dataset's scale and diversity across the selected tasks, are summarized in Table~\ref{tab:dataset_statistics_finetuned}.


\begin{table}[t]
\centering
\caption{Detailed HHoI Dataset Statistics by Interaction Task. The column abbreviations are as follows: \textbf{Sam.}: Samples; \textbf{Seq.}: Total sequences of frames; \textbf{Dura.}: Total duration in seconds; \textbf{Avg Fra.}: Average frames per sample; \textbf{Avg Sec.}: Average seconds per sample.}
\label{tab:dataset_statistics_finetuned} 
\footnotesize
\resizebox{\columnwidth}{!}{%
\begin{tabularx}{\linewidth}{l *{5}{>{\centering\arraybackslash}X}}
\toprule 
\textbf{Task} & \textbf{Sam.} & \textbf{Seq.} & \textbf{Dura.} & \textbf{Avg Fra.} & \textbf{Avg Sec.} \\
\midrule 
Hug & 226 & 292,873 & 5,857.5 & 1,296 & 25.9 \\
Handshake & 259 & 296,414 & 5,928.3 & 1,144 & 22.9 \\
High-Five & 219 & 176,482 & 3,529.6 & 805 & 16.1 \\
Bend & 200 & 176,024 & 3,520.5 & 880 & 17.6 \\
Fly-Kiss & 177 & 158,412 & 3,168.2 & 895 & 17.9 \\
Wave & 160 & 149,118 & 2,982.4 & 932 & 18.6 \\
\midrule 
\textbf{Total} & \textbf{1,241} & \textbf{1,249,323} & \textbf{24,986.5} & \textbf{1,007} & \textbf{20.1} \\
\bottomrule 
\end{tabularx}
}
\end{table}

\begin{table}[t]
\centering
\caption{\textbf{Symbol table (core).} In this table, $(\tau)$ denotes the \textbf{contact distance threshold} only; task-detection thresholds are denoted $(\tau_{\cdot})$ and are summarized in Table~S6.}
\resizebox{\columnwidth}{!}{%
\begin{tabular}{l l}
\toprule
Symbol & Meaning \\
\midrule
$\mathbf{M}_X=\{\mathbf{q}^X_t\}_{t=1}^{T}$ & Motion sequence of agent $X$ \\
$\mathcal{J}_X$ & Joint set of agent $X$ \\
$\mathcal{J}_t(X,j)\in\mathbb{R}^3$ & 3D joint position via FK from $\mathbf{q}^X_t$ \\
$\text{Reshaped}(H_s)$ & Morphology-aligned skeleton of $H_s$ \\
$\mathbf{P}_{\text{task}}$ & Task-specific human--robot keypoint pairs \\
$\Delta t$ & Sampling period (0.02\,s, i.e., 50\,Hz) \\
$\tau$ & Contact distance threshold ($0.2/0.35/0.5$ m) \\
${w}_{\text{phase\_KL}}$ & Weight for the phase/transition KL term ($\mathcal{L}_{\text{trans}}$) \\ 
\textbf{PAIR} & Physics-Aware Interaction Retargeting \\
\textbf{D-STAR} & Decoupled Spatio-Temporal Action Reasoner \\
\bottomrule
\end{tabular}
\vspace{-10mm}
\label{tab:symbols}
}
\end{table}

\noindent\textbf{Temporal and Spatial Richness:} The final dataset spans approximately 6.9 hours of interaction data (at 50\,Hz), with average sample durations ranging from 16.1 seconds for compact gestures to 25.9 seconds for complex, contact-rich tasks like hugs. This temporal depth, combined with the tracking of key body joints, provides a rich source of spatio-temporal data essential for training our multi-scale policy. The balanced distribution across tasks ensures robust training and evaluation across all targeted interaction categories.

\section{Experimental Protocols and Evaluation Metrics}
\label{sec:sup_protocols}\label{app:exp}

In this section, we provide a detailed account of the protocols and metrics used to validate our framework at both the data and policy levels. The descriptions are designed to be comprehensive, ensuring full reproducibility of our experimental results and the claims made in the main paper.

\subsection{Unified Notation \& Units}
\label{app:c1-notation}\label{app:notation}

\paragraph{Sampling and units.}
All sequences are sampled at $\Delta t=0.02$\,s (50\,Hz). Distances are in meters; angles in radians; jerk on positional trajectories is in m/s$^3$.

\paragraph{Motion variables.}
For agent $X\in\{H_s,H_p,R\}$ (source human, partner human, robot), a motion sequence is $\mathbf{M}_X=\{\mathbf{q}^X_t\}_{t=1}^{T}$ with generalized coordinates $\mathbf{q}^X_t\in\mathbb{R}^{D_X}$. Let $\mathcal{J}_X$ be the joint set of $X$. The 3D position of joint $j\in\mathcal{J}_X$ is $\mathcal{J}_t(X,j)\in\mathbb{R}^3$, obtained by forward kinematics from $\mathbf{q}^X_t$.

\paragraph{Morphology alignment.}
$\text{Reshaped}(H_s)$ denotes the morphology-aligned human skeleton constructed by pelvis-frame alignment, isotropic bone-length scaling, and a fixed joint correspondence derived from the robot--SMPL mapping. In practice, we obtain the reshaped SMPL parameters by optimizing SMPL shape $\boldsymbol{\beta}$ and a global scale $s$ to fit robot bone lengths (see Sec.~\ref{sec:sup_retargeting_details}); we refer to the resulting kinematic chain as $\text{Reshaped}(H_s)$.

\paragraph{Task keypoint pairs.}
$\mathbf{P}_{\text{task}}$ denotes the task-specific set of human--robot keypoint pairs referenced by contact-related definitions (see Sec.~\ref{sec:sup_retargeting_metrics}).

\subsection{Motion Retargeting Evaluation}
\label{sec:sup_retargeting_eval}\label{app:retarget_eval}

To rigorously assess the quality of \textbf{PAIR}, we employed a comprehensive suite of 18 metrics spanning physical consistency, contact preservation, plausibility, and smoothness. 

\subsubsection{Evaluation Metrics}
\label{sec:sup_retargeting_metrics}

\paragraph{Physical Consistency Metrics.}
These metrics evaluate the overall geometric and structural fidelity of the retargeted motion.
\begin{itemize}
    \item \textbf{Joint Position Error (JPE) $\downarrow$}: The mean L2 distance between robot joints and the morphologically reshaped source human joints $\text{Reshaped}(H_s)$ (Sec.~\ref{sec:sup_retargeting_details}), measuring kinematic similarity.
    \item \textbf{Average Workspace Distance (AWD) $\downarrow$}: The mean absolute difference between the full $N\!\times\!N$ pairwise distance matrices built from a fixed set of interaction joints (head, shoulders, elbows, wrists). We compare the original human--human interaction (initiator vs. responder) with the retargeted human--robot interaction (initiator vs. robot), capturing holistic spatial-relationship preservation.
\end{itemize}

\paragraph{Multi-Threshold Contact Detection Metrics.}
To evaluate contact preservation across varying interaction types, we define three distance thresholds and compute standard classification metrics for each.
\begin{itemize}
    \item \textbf{Thresholds}: 0.2\,m, 0.35\,m, and 0.5\,m.
    \item \textbf{Protocol}: For each frame we classify contact for the two hand--hand pairs using optimal left/right matching; metrics are micro-averaged over frames$\times$hands. Unless otherwise stated, we do not enforce a minimum contact duration.
    \item \textbf{Metrics}: For each threshold, we compute \textbf{Precision $\uparrow$}, \textbf{Recall $\uparrow$}, \textbf{F1-Score $\uparrow$}, and \textbf{Accuracy $\uparrow$} to quantify how well the retargeted motion preserves the original contact semantics.
\end{itemize}

\paragraph{Physical Plausibility Metrics.}
These metrics assess whether the generated motion is natural and within realistic biomechanical limits.
\begin{itemize}
    \item \textbf{Large Angle Ratio $\downarrow$}: The percentage of frames where joint-angle magnitudes (axis--angle) exceed $0.5$\,rad, flagging unnatural poses.
    \item \textbf{Angle Standard Deviation $\downarrow$}: The standard deviation of joint-angle magnitudes (axis--angle) aggregated over joints and frames, indicating pose distribution consistency.
\end{itemize}

\paragraph{Motion Smoothness Metrics.}
These metrics quantify the temporal coherence and fluidity of the motion.
\begin{itemize}
    \item \textbf{Jerk Mean $\downarrow$ (m/s$^3$)}: The average magnitude of the third temporal derivative of 3D joint positions, computed via third-order finite differences at the dataset frame rate (50\,Hz), penalizing high-frequency jitter.
    \item \textbf{Jerk Standard Deviation $\downarrow$ (m/s$^3$)}: The standard deviation of the same jerk magnitudes, indicating temporal stability.
\end{itemize}

\subsection{Interaction Policy Evaluation}
\label{sec:sup_policy_eval}\label{app:policy_eval}

\subsubsection{Task Success Criteria and Detection Algorithms}
\label{sec:sup_detection_algos}
To quantitatively evaluate policy performance, we designed specific success criteria for each of the six interaction tasks. The core logic for each detection algorithm is outlined below (Algorithms~\ref{alg:hug_detection_full}--\ref{alg:flyingkiss_detection_full}), providing a transparent basis for our results.

\subsubsection*{Hug}
A hug is detected if the human's hands make sustained and appropriate contact with the robot's torso or shoulders, captured via three distinct modes.
Algorithm~\ref{alg:hug_detection_full} details the full detection logic.

\subsubsection*{Handshake}
A successful handshake is verified through both sustained hand contact and appropriate interpersonal distance.
Algorithm~\ref{alg:handshake_detection_full} provides the handshake detection procedure.

\subsubsection*{High-Five}
Success requires a brief, decisive contact between human and robot hands at an appropriate height.
Algorithm~\ref{alg:highfive_detection_full} outlines the high-five detector.

\subsubsection*{Wave}
A wave is identified by characteristic hand motion patterns, including sufficient amplitude and clear direction changes.
Algorithm~\ref{alg:wave_detection_full} describes the full wave detection pipeline.

\subsubsection*{Bend}
A bend is robustly measured by the forward inclination angle of the head-root vector relative to the vertical axis.
Algorithm~\ref{alg:bend_detection_full} formalizes the bend detection criteria.

\subsubsection*{Fly-Kiss}
This gesture is detected via its characteristic two-phase sequence: hand-to-head proximity followed by a forward projection motion.
Algorithm~\ref{alg:flyingkiss_detection_full} summarizes the fly-kiss detector.

\subsubsection{Hyperparameter Justification}
The hyperparameters for these detection algorithms, detailed in Table~\ref{tab:detection_hyperparameters}, were empirically determined by tuning on a randomly sampled validation subset of 50 annotated interaction sequences drawn from the full dataset. The values were selected to maximize the alignment between our automated metrics and the consensus of three human evaluators, ensuring a robust and meaningful evaluation that reflects human judgment of interaction success.

\section{Dataset Retargeting for Sim-to-Real}
\label{app:retarget}\label{app:sim2real}

\textbf{Protocol.} We keep \textit{Act} unchanged. \textit{Preparation} is standardized to a neutral upper-body posture (both hands naturally placed in front of the root on the left/right sides), with identical lower-body motion; the root is retargeted to milder yaw and simpler walking. The last 16 frames bilinearly transition into the first frame of the \textit{Act}. \textit{Follow-up} becomes stationary while the upper body bilinearly returns over 16 frames to the same neutral posture.

\textbf{Pseudo-code.} Algorithm~\ref{alg:retarget} summarizes the retargeting routine.

\section{Observation \& Action Specification}
\label{app:spec}

\noindent\textbf{Observation and action specification.} We use human root translation (3D) and SMPL joint positions (72D) as the human observation channels, combined with robot proprioception. The reference action is 36-D (29-DoF joint targets + root translation/orientation). Table~\ref{tab:obs} lists the exact channels and dimensionalities.

\section{Temporal Scheduler: Densify and Fuse (within D-STAR)} 
\label{app:scheduler}\label{app:schedule}

\noindent\textbf{Diffusion details.} The diffusion-based planning head is trained with a DDIM scheduler~\cite{song2020denoising} using 50 training time steps and 10 inference steps with a squared-cosine schedule. We use group normalization throughout the denoiser and adopt standard sinusoidal timestep embeddings.

\noindent\textbf{Runtime pipeline.} The policy emits 5 anchors per invocation, covering 0--2.0\,s with a 0.5\,s step; the policy runs at 5\,Hz. Anchors are densified to 50\,Hz and bilinearly fused across overlapping calls to ensure temporal continuity; zero-order hold is applied if updates are delayed and root-velocity bounds are temporarily tightened.

\textbf{Densification (per call at $t_k$).} Algorithm~\ref{alg:densify} interpolates each 5-anchor policy output to a 50\,Hz reference trajectory.

\textbf{Cross-call fusion (between $t_k$ and $t_{k+1}$, 5\,Hz).} Algorithm~\ref{alg:schedule} bilinearly fuses overlapping dense segments to ensure continuity.

\begin{table*}[!t]
\centering
\caption{Detection Algorithm Hyperparameters}
\label{tab:detection_hyperparameters}\label{tab:thresholds}
\small
\setlength{\tabcolsep}{4pt}
\begin{tabularx}{\linewidth}{l l >{\centering\arraybackslash}X l}
\hline
\textbf{Task} & \textbf{Parameter} & \textbf{Value} & \textbf{Description} \\
\hline
\multirow{4}{*}{Hug} 
& $\tau_{hand\_dist}$ & 0.5m & Maximum distance between hands for double embrace \\
& $\tau_{body}$ & 0.45m & Hand to robot torso threshold \\
& $\tau_{shoulder}$ & 0.4m & Hand to shoulder distance for shoulder embrace \\
& $\tau_{min\_frames}$ & 3 & Minimum consecutive frames for a successful embrace \\
\hline
\multirow{4}{*}{High-Five} 
& $\tau_{contact}$ & 0.4m & Maximum hand-to-hand distance for contact \\
& $\tau_{min\_approach}$ & 1 & Minimum frames of valid contact approach \\
& $\tau_{max\_sustained}$ & 1000 & Maximum frames of sustained contact (to rule out holding) \\
& $\tau_{height}$ & 0.3m & Minimum hand height above human's root \\
\hline
\multirow{6}{*}{Handshake} 
& $\tau_{contact}$ & 0.3m & Maximum hand-to-hand contact distance \\
& $\tau_{min\_contact}$ & 10 & Minimum consecutive frames of sustained contact \\
& $\tau_{min\_dist}$ & 0.4m & Minimum distance between agent roots \\
& $\tau_{max\_dist}$ & 1.5m & Maximum distance between agent roots \\
& $\tau_{std\_contact}$ & 0.15m & Upper bound of contact-distance standard deviation (stability) \\
& $\tau_{mean\_contact}$ & 0.45m & Upper bound of contact-distance mean (stability) \\
\hline
\multirow{5}{*}{Wave} 
& $\tau_{motion\_dist}$ & 0.5m & Minimum total distance traveled by the waving hand \\
& $\tau_{min\_changes}$ & 3 & Minimum number of significant direction changes \\
& $\tau_{amplitude}$ & 0.3m & Minimum motion amplitude (peak-to-peak) \\
& $\tau_{angle}$ & 0.785 rad & Angle threshold for detecting a direction change ($\pi/4$) \\
& $\tau_{height}$ & 0.45m & Minimum hand height above human's root \\
\hline
\multirow{2}{*}{Bend} 
& $\tau_{min\_angle}$ & 0.349 rad & Minimum bend angle from vertical ($\pi/6$) \\
& $\tau_{min\_frames}$ & 5 & Minimum consecutive frames angle is above threshold \\
\hline
\multirow{3}{*}{Fly-Kiss} 
& $\tau_{hand2head}$ & 0.3m & Maximum distance from hand to head for Phase 1 \\
& $\tau_{forward\_thresh}$ & 0.1m & Minimum total forward motion projected towards partner \\
& $\tau_{min\_forward}$ & 5 & Minimum consecutive frames of forward hand motion in Phase 2 \\
\hline
\end{tabularx}
\end{table*}

\begin{algorithm*}[t]
\caption{Hug Detection with Multi-Modal Embrace Recognition}
\label{alg:hug_detection_full}
\begin{algorithmic}[1]
\REQUIRE Human joints $\mathbf{J}_H$, robot torso position $\mathbf{p}_R$
\ENSURE Success flag $success_{hug}$
\STATE Extract hand positions: $\mathbf{h}_L, \mathbf{h}_R \gets \mathbf{J}_H[joints_{hand}]$
\STATE $double\_embrace \gets \text{DetectDoubleEmbrace}(\mathbf{h}_L, \mathbf{h}_R, \mathbf{p}_R)$
\STATE $single\_embrace \gets \text{DetectSingleEmbrace}(\mathbf{h}_L, \mathbf{h}_R, \mathbf{p}_R)$
\STATE $shoulder\_embrace \gets \text{DetectShoulderEmbrace}(\mathbf{h}_L, \mathbf{h}_R, \mathbf{p}_R)$
\STATE \textbf{return} $double\_embrace \lor single\_embrace \lor shoulder\_embrace$

\vspace{0.5em}
\PROCEDURE{DetectDoubleEmbrace}{$\mathbf{h}_L, \mathbf{h}_R, \mathbf{p}_R$}
\STATE $d_{hands} \gets \|\mathbf{h}_L - \mathbf{h}_R\|_2$ \COMMENT{Distance between hands}
\STATE $d_{L2robot} \gets \|\mathbf{h}_L - \mathbf{p}_R\|_2$, $d_{R2robot} \gets \|\mathbf{h}_R - \mathbf{p}_R\|_2$
\STATE $hands\_close \gets d_{hands} < \tau_{hand\_dist}$
\STATE $hands\_near\_robot \gets (d_{L2robot} < \tau_{body}) \land (d_{R2robot} < \tau_{body})$
\STATE $embrace\_frames \gets hands\_close \land hands\_near\_robot$
\STATE \textbf{return} $\text{MaxConsecutiveFrames}(embrace\_frames) \geq \tau_{min\_frames}$
\ENDPROCEDURE
\end{algorithmic}
\vspace{-1mm}
\end{algorithm*}

\begin{algorithm*}[t]
\caption{Handshake Detection with Spatial Validation}
\label{alg:handshake_detection_full}
\begin{algorithmic}[1]
\REQUIRE Human hands $\mathbf{h}_{L,H}, \mathbf{h}_{R,H}$, robot hands $\mathbf{h}_{L,R}, \mathbf{h}_{R,R}$, human root $\mathbf{r}_H$
\ENSURE Success flag $success_{handshake}$
\STATE $d_{min} \gets \text{ComputeMinHandDist}(\mathbf{h}_{\text{L,H}}, \mathbf{h}_{\text{R,H}}, \mathbf{h}_{\text{L,R}}, \mathbf{h}_{\text{R,R}})$
\STATE $contact\_phase \gets \text{DetectContactPhase}(d_{min}, \mathbf{r}_H)$
\STATE $stability\_phase \gets \text{DetectStabilityPhase}(d_{min}, \mathbf{r}_H)$
\STATE \textbf{return} $contact\_phase \land stability\_phase$

\PROCEDURE{DetectContactPhase}{$d_{min}, \mathbf{r}_H$}
\STATE $contact\_frames \gets d_{min} < \tau_{contact}$
\STATE Validate each contact frame with root distance constraint:
\FOR{$t \in contact\_frames$}
    \STATE $\mathbf{h}_{active} \gets \text{GetActiveHand}(t)$ \COMMENT{Hand with min dist}
    \STATE $d_{root} \gets \|\mathbf{h}_{active} - \mathbf{r}_H[t]\|_2$
    \STATE $contact\_frames[t] \gets (d_{root} \geq \tau_{min\_dist}) \land (d_{root} \leq \tau_{max\_dist})$
\ENDFOR
\STATE $valid\_contact\_count \gets \sum contact\_frames$
\STATE $max\_consecutive \gets \text{MaxConsecutiveFrames}(contact\_frames)$
\STATE \textbf{return} $(valid\_contact\_count \geq \tau_{min\_contact}) \land (max\_consecutive \geq \tau_{min\_contact})$
\ENDPROCEDURE

\vspace{0.25em}
\PROCEDURE{DetectStabilityPhase}{$d_{min}, \mathbf{r}_H$}
\STATE $contact\_frames \gets d_{min} < \tau_{contact}$
\STATE $d\_seq \gets d_{min}[contact\_frames]$
\STATE $stability\_std \gets \text{RollingStd}(d\_seq)$
\STATE $stability\_mean \gets \text{RollingMean}(d\_seq)$
\STATE \textbf{return} $(\text{Std}(d\_seq) < \tau_{std\_contact}) \land (\text{Mean}(d\_seq) < \tau_{mean\_contact})$
\ENDPROCEDURE
\end{algorithmic}
\end{algorithm*}

\begin{algorithm*}[t]
\caption{High-Five Detection with Height Validation}
\label{alg:highfive_detection_full}
\begin{algorithmic}[1]
\REQUIRE Human hands $\mathbf{h}_{L,H}, \mathbf{h}_{R,H}$, robot hands $\mathbf{h}_{L,R}, \mathbf{h}_{R,R}$, human root $\mathbf{r}_H$
\ENSURE Success flag $success_{highfive}$
\STATE Compute all hand-to-hand distances:
\STATE $d_{LL} \gets \|\mathbf{h}_{L,H} - \mathbf{h}_{L,R}\|_2$, $d_{LR} \gets \|\mathbf{h}_{L,H} - \mathbf{h}_{R,R}\|_2$
\STATE $d_{RL} \gets \|\mathbf{h}_{R,H} - \mathbf{h}_{L,R}\|_2$, $d_{RR} \gets \|\mathbf{h}_{R,H} - \mathbf{h}_{R,R}\|_2$
\STATE $d_{min} \gets \min(d_{LL}, d_{LR}, d_{RL}, d_{RR})$
\STATE $close\_frames \gets d_{min} < \tau_{contact}$
\IF{height\_validation\_enabled}
    \STATE $valid\_frames \gets \text{ValHandHeight}(close\_frames, \mathbf{h}_{\text{L,H}}, \mathbf{h}_{\text{R,H}}, \mathbf{h}_{\text{L,R}}, \mathbf{h}_{\text{R,R}}, \mathbf{r}_\text{H})$
    \STATE $close\_frames \gets close\_frames \land valid\_frames$
\ENDIF
\STATE $contact\_count \gets \sum close\_frames$
\STATE $consecutive\_count \gets \text{MaxConsecutiveFrames}(close\_frames)$
\STATE \textbf{return} $(contact\_count \geq \tau_{min\_approach}) \land (consecutive\_count \leq \tau_{max\_sustained})$
\end{algorithmic}
\end{algorithm*}

\begin{algorithm*}[t]
\caption{Wave Detection with Motion Pattern Analysis}
\label{alg:wave_detection_full}
\begin{algorithmic}[1]
\REQUIRE Human hands $\mathbf{h}_{L,H}, \mathbf{h}_{R,H}$, human root $\mathbf{r}_H$ 
\ENSURE Success flag $success_{wave}$
\STATE Select active hand based on motion intensity.
\STATE $motion_L \gets \text{CalculateMotionIntensity}(\mathbf{h}_{L,H})$
\STATE $motion_R \gets \text{CalculateMotionIntensity}(\mathbf{h}_{R,H})$
\STATE $\mathbf{h}_{active} \gets \mathbf{h}_{L,H}$ if $motion_L.total > motion_R.total$ else $\mathbf{h}_{R,H}$
\STATE $motion\_sufficient \gets motion.total > \tau_{motion\_dist}$
\STATE $direction\_changes \gets \text{DetectDirectionChanges}(\mathbf{h}_{active})$
\STATE $amplitude\_check \gets \text{DetectAmplitude}(\mathbf{h}_{active})$
\STATE $height\_check \gets \text{DetectHandHeight}(\mathbf{h}_{L,H}, \mathbf{h}_{R,H}, \mathbf{r}_H)$
\STATE \textbf{return} $motion\_sufficient \land direction\_changes \land amplitude\_check \land height\_check$

\vspace{0.5em}
\PROCEDURE{DetectDirectionChanges}{$\mathbf{h}_{active}$}
\STATE Compute position changes: $\Delta \mathbf{p} \gets \text{diff}(\mathbf{h}_{active})$
\STATE Compute motion angles: $\theta \gets \arctan2(\Delta \mathbf{p}[:, 1], \Delta \mathbf{p}[:, 0])$
\STATE $direction\_changes \gets 0$
\FOR{$i = 1$ to $\text{len}(\theta)-1$}
    \STATE $\Delta \theta \gets |\theta[i] - \theta[i-1]|$
    \IF{$\Delta \theta > \pi$} $\Delta \theta \gets 2\pi - \Delta \theta$ \ENDIF
    \IF{$\Delta \theta > \tau_{angle}$ and $\text{CheckDisplacement}(\mathbf{h}_{active}, i)$}
        \STATE $direction\_changes \gets direction\_changes + 1$
    \ENDIF
\ENDFOR
\STATE \textbf{return} $direction\_changes \geq \tau_{min\_changes}$
\ENDPROCEDURE
\end{algorithmic}
\end{algorithm*}

\begin{algorithm*}[t]
\caption{Bend Detection via Head-Root Angular Analysis}
\label{alg:bend_detection_full}
\begin{algorithmic}[1]
\REQUIRE Human head $\mathbf{h}_{head}$, human root $\mathbf{r}_H$
\ENSURE Success flag $success_{bend}$
\STATE Compute head-root vectors: $\mathbf{v}_{hr} \gets \mathbf{h}_{head} - \mathbf{r}_H$
\STATE Define vertical axis: $\mathbf{z}_{axis} \gets [0, 0, 1]$
\STATE Compute bend angles: $\theta_{bend} \gets \text{VectorAngle}(\mathbf{v}_{hr}, \mathbf{z}_{axis})$
\STATE $max\_angle \gets \max(\theta_{bend})$
\STATE $frames\_above\_threshold \gets \sum(\theta_{bend} \geq \tau_{min\_angle})$
\STATE $angle\_condition \gets max\_angle \geq \tau_{min\_angle}$
\STATE $stability\_condition \gets frames\_above\_threshold \geq \tau_{min\_frames}$
\STATE \textbf{return} $angle\_condition \land stability\_condition$
\end{algorithmic}
\end{algorithm*}

\begin{algorithm*}[t]
\caption{Fly-Kiss Detection with Motion Projection}
\label{alg:flyingkiss_detection_full}
\begin{algorithmic}[1]
\REQUIRE Human hands $\mathbf{h}_{L,H}, \mathbf{h}_{R,H}$, head $\mathbf{h}_{head}$, roots $\mathbf{r}_H, \mathbf{r}_R$
\ENSURE Success flag $success_{flyingkiss}$
\STATE Select active hand closest to head.
\STATE $d_{L2head} \gets \min(\|\mathbf{h}_{L,H} - \mathbf{h}_{head}\|_2)$
\STATE $d_{R2head} \gets \min(\|\mathbf{h}_{R,H} - \mathbf{h}_{head}\|_2)$
\STATE $\mathbf{h}_{active}, d_{min} \gets (\mathbf{h}_{L,H}, d_{L2head})$ if $d_{L2head} \leq d_{R2head}$ else $(\mathbf{h}_{R,H}, d_{R2head})$
\STATE $condition\_A \gets d_{min} \leq \tau_{hand2head}$
\STATE Compute forward direction: $\mathbf{dir} \gets \text{normalize}(\text{mean}(\mathbf{r}_R) - \text{mean}(\mathbf{r}_H))$
\STATE Compute hand motion: $\Delta \mathbf{h} \gets \text{diff}(\mathbf{h}_{active})$
\STATE Forward projections: $proj \gets \Delta \mathbf{h} \cdot \mathbf{dir}$
\STATE $forward\_frames \gets proj > 0$
\STATE $consecutive\_forward \gets \text{MaxConsecutiveFrames}(forward\_frames)$
\STATE $total\_forward\_motion \gets \sum(proj[proj > 0])$
\STATE $condition\_B \gets (consecutive\_forward \geq \tau_{min\_forward}) \land (total\_forward\_motion \geq \tau_{forward\_thresh})$
\STATE \textbf{return} $condition\_A \land condition\_B$
\end{algorithmic}
\end{algorithm*}

\begin{algorithm*}[t]
\caption{Retarget(Preparation, Act, Follow-up)}
\label{alg:retarget}
\begin{algorithmic}[1]
\STATE $\text{Prep}\leftarrow$ neutralize\_upper\_body(Preparation)
\STATE $\text{Prep.root}\leftarrow$ retarget\_root(Preparation.root, mild\_yaw, simple\_walk)
\STATE $\text{Act}\leftarrow$ Act \hfill
\STATE $\text{Follow}\leftarrow$ make\_stationary(Follow\_up)
\STATE $\text{Follow.upper}\leftarrow$ return\_to\_neutral(Follow\_up.upper)
\STATE // 16-frame bilinear ramps
\STATE $\text{Prep}[-16:]\leftarrow$ bilinear\_ramp($\text{Prep}[-16:], \text{Act}[0]$)
\STATE $\text{Follow}[:16]\leftarrow$ bilinear\_ramp($\text{Act}[-1], \text{Follow}[:16]$)
\RETURN concatenation($\text{Prep}, \text{Act}, \text{Follow}$)
\end{algorithmic}
\end{algorithm*}

\begin{algorithm*}[t]
\caption{DensifyAndTrack(5 anchors $\rightarrow$ 100-frame reference)}\label{alg:densify}
\begin{algorithmic}[1]
\STATE anchors $\{(t_k + 0.0, \mathbf{q}_0), (t_k + 0.5, \mathbf{q}_{0.5}), \ldots, (t_k + 2.0, \mathbf{q}_{2.0})\}$
\STATE grid $\mathcal{T} = \{t_k + i/50 \mid i=0,\ldots,99\}$
\FOR{$t \in \mathcal{T}$}
  \STATE $(\tau_1,\tau_2) \leftarrow$ nearest anchor times s.t. $\tau_1 \le t \le \tau_2$
  \STATE $\hat{\mathbf{q}}(t) \leftarrow$ \textbf{bilinear\_interp} in (time $\times$ joint/root) between $(\tau_1,\mathbf{q}_{\tau_1}), (\tau_2,\mathbf{q}_{\tau_2})$
\ENDFOR
\RETURN $\{\hat{\mathbf{q}}(t)\}_{t\in\mathcal{T}}$
\end{algorithmic}
\end{algorithm*}

\begin{algorithm*}[t]
\caption{FuseOverlaps($\hat{\mathbf{q}}_k$, $\hat{\mathbf{q}}_{k+1}$)}\label{alg:schedule}
\begin{algorithmic}[1]
\STATE let $\mathcal{T}_{\text{overlap}} = \mathcal{T}_k \cap \mathcal{T}_{k+1}$ on the 50\,Hz grid
\FOR{$t \in \mathcal{T}_{\text{overlap}}$}
  \STATE $\tilde{\mathbf{q}}(t) \leftarrow$ \textbf{bilinear\_interp\_time}$(\hat{\mathbf{q}}_k(t), \hat{\mathbf{q}}_{k+1}(t), t)$
\ENDFOR
\STATE if timeout: $\tilde{\mathbf{q}}(t) \leftarrow$ hold-last and tighten root-velocity bounds
\RETURN fused dense segment $\tilde{\mathbf{q}}(t)$
\end{algorithmic}
\end{algorithm*}

\end{document}